\documentclass[]{externalized}
\usepackage[T1]{fontenc}
\usepackage[utf8]{inputenc}
\usepackage{microtype}
\usepackage{inconsolata}
\usepackage{graphicx}
\usepackage{hyperref}
\usepackage{url}
\usepackage{svg}
\usepackage{booktabs}
\usepackage{amsfonts}
\usepackage{amsmath}
\usepackage{amssymb}
\usepackage{amsthm}
\usepackage{mathtools}
\usepackage{enumitem}
\usepackage{multirow}
\usepackage{subcaption}
\usepackage{xcolor}
\usepackage{listings}
\usepackage{natbib}
\usepackage{multicol}
\usepackage{pifont}
\usepackage{xspace}
\usepackage{wrapfig}
\AtBeginDocument{%
  \providecommand\BibTeX{{%
    \normalfont B\kern-0.5em{\scshape i\kern-0.25em b}\kern-0.8em\TeX}}}

\makeatletter
\DeclareRobustCommand\onedot{\futurelet\@let@token\@onedot}
\def\@onedot{\ifx\@let@token.\else.\null\fi}

\makeatother

\definecolor{codegreen}{rgb}{0,0.6,0}
\definecolor{codegray}{rgb}{0.5,0.5,0.5}
\definecolor{codepurple}{rgb}{0.58,0,0.82}
\definecolor{backcolour}{rgb}{0.95,0.95,0.92}

\lstdefinestyle{mystyle}{
  backgroundcolor=\color{backcolour},
  commentstyle=\color{codegreen},
  keywordstyle=\color{magenta},
  numberstyle=\tiny\color{codegray},
  stringstyle=\color{codepurple},
  basicstyle=\ttfamily\footnotesize,
  breakatwhitespace=false,
  breaklines=true,
  captionpos=b,
  keepspaces=true,
  numbers=left,
  numbersep=5pt,
  showspaces=false,
  showstringspaces=false,
  showtabs=false,
  tabsize=2
}

\usepackage[most]{tcolorbox}
\usepackage{xcolor}
\definecolor{myblue}{RGB}{62,94,148}
\usepackage{graphicx}
\usepackage{eso-pic}
\usepackage{tabularx}
\usepackage{booktabs}
\usepackage{array}
\usepackage[table]{xcolor}
\usepackage{hyperref}
\definecolor{myorange}{HTML}{FD6000}
\definecolor{complexityblue}{HTML}{2963C1}
\definecolor{diversitygreen}{HTML}{3A5F21}

\title{What Makes Good Agentic Data? An 
\textcolor[HTML]{FD6000}{A}\textcolor[HTML]{2963C1}{C}\textcolor[HTML]{3A5F21}{E}
 Lens on \\Data Generation for LLM Agents}
\author[1]{Xingshan Zeng}
\author[2]{Zishan Xu}
\author[2]{Boju Zhang}
\author[3]{Yuzhou Wu}
\author[4]{Lingzhi Wang}
\author[2]{Jianghao Lin}
\author[5]{Liangyou Li}
\author[]{Yasheng Wang}
\author[1]{Lifeng Shang}
\author[1]{Xin Jiang}
\author[2]{Weinan Zhang}
\author[2]{Yong Yu}
\author[1]{Qun Liu}
\author[2,\dag]{Weiwen Liu}

\affiliation[1]{Huawei Technologies Co., Ltd}
\affiliation[2]{Shanghai Jiao Tong University}
\affiliation[3]{Northwestern University, Chicago, IL} 
\affiliation[4]{Harbin Institute of Technology, Shenzhen}
\affiliation[5]{Shenzhen Loop Area Institute}
\contribution[\dag]{Corresponding author}
\correspondence{\email{zeng.xingshan@huawei.com}, \email{wwliu@sjtu.edu.cn}}

\AddToShipoutPictureBG*{%
  \AtPageLowerLeft{%
    \raisebox{1.2cm}{%
      \makebox[\paperwidth][c]{%
        \includegraphics[
          width=0.87\paperwidth
        ]{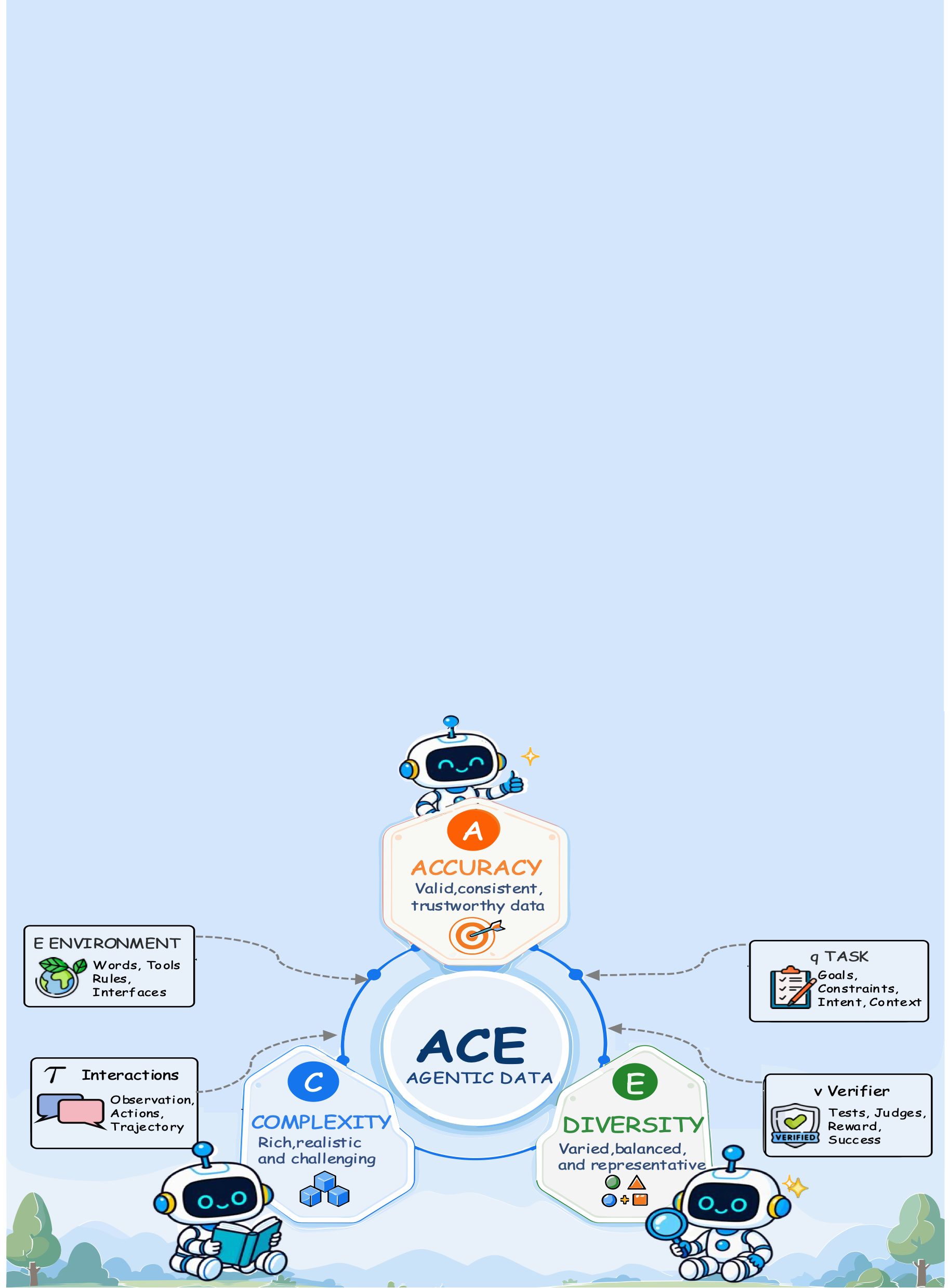}%
      }%
    }%
  }%
}
\abstract{
Large language model agents increasingly rely on generated interaction data to learn how to interact with external environments. Unlike conventional instruction synthesis, agentic data generation must maintain consistency among environments, tasks, interactions, and success signals while producing experience that is useful rather than merely abundant. Existing work spans many agent domains, but domain-centered organization and heterogeneous evaluation often obscure common generation mechanisms and conflate candidate construction with verification and selection.
This work develops a two-level framework for the field. First, we represent agentic data as a common factorized object $(E,q,\tau,v)$, comprising an environment specification, task signal, interaction realization, and optional verifier. We organize generation paradigms by their primary anchor and dependency structure, covering general forward and reverse pipelines. Second, we formulate generation as constrained distribution design through the \emph{
\textcolor[HTML]{FD6000}{\textbf{A}}ccuracy--
\textcolor[HTML]{2963C1}{\textbf{C}}omplexity--
div\textcolor[HTML]{3A5F21}{\textbf{E}}rsity
}
(\textcolor[HTML]{FD6000}{\textbf{A}}\textcolor[HTML]{2963C1}{\textbf{C}}\textcolor[HTML]{3A5F21}{\textbf{E}}) lens. Accuracy establishes the feasible support of grounded and internally consistent data. Within this support, Complexity places learning mass relative to the capability of a declared learner and execution configuration, while divErsity controls coverage and redundancy across environments, tasks, and interaction behaviors.
Using this framework, we explore how prior work verifies generated experience, constructs and calibrates difficulty, and expands behavioral coverage. The literature reveals a shift toward execution-grounded accuracy, learner-relative complexity, and diversity beyond surface variation or dataset size. We further discuss broader directions and emerging trends in agentic data generation through the ACE lens, including their implications for scaling, data sources, training regimes and adaptive learning. Overall, the central challenge is not simply to generate more data, but to continually allocate valid, informative, and non-redundant experience as agents and environments evolve.
}

\date{\today}

\begin{document}
\maketitle
\newpage
\tableofcontents
\newpage

% Major sections live under text/ so the manuscript stays modular.
\section{Introduction}
\label{sec:introduction}

Large language model (LLM) agents are increasingly expected to act rather than only respond: they invoke tools, operate systems, search information, modify files, and interact with simulated or physical worlds~\citep{qinToolLLMFacilitatingLarge2023,xieOSWorldBenchmarking2024,pan2024training,zhengDeepResearcher2025,wangRoboGenTowardsUnleashing2023}. Learning these behaviors requires experience that connects decisions to observations and state changes, often over multiple turns~\citep{prabhakarAPIGenMTAgenticPipeline2025,dongAgentWorldScalingRealWorld2026}. Such interaction data are expensive to collect manually, and difficult to verify and expand. Agentic data generation has therefore become a central means of scaling both agent training and evaluation~\citep{mitraAgentInstructGenerativeTeaching2024,liuAPIGenAutomatedPipeline2024,songEnvScalerScalingToolInteractive2026}. The central challenge, however, is not merely to generate more data, but to ensure that generated experience is accurate, appropriately complex for the learner, and sufficiently diverse.

Agentic data generation is not simply instruction generation followed by response sampling. A useful sample must connect an actionable environment, a grounded task, and an interaction in which actions produce valid observations and state changes. 
% It may also require a verifier or reward interface that recognizes the intended process or outcome. 
% A fluent trajectory can still be unusable when its task is infeasible, its tools are inconsistent, its observations are fabricated, or its success signal wrongly rewards. 
Recent pipelines consequently generate not only instructions and demonstrations, but also tool ecosystems, executable environments, stateful tasks, rollout feedback, and verification procedures~\citep{liuAPIGenAutomatedPipeline2024,liuToolACEWinningPoints2025,xuTOUCANSynthesizing15M2025,songEnvScalerScalingToolInteractive2026,xuEnvFactoryScalingToolUse2026}. We use \emph{Accuracy--Complexity--divErsity} (ACE) to name the three corresponding requirements: whether generated experience is grounded and valid, whether it presents a useful challenge for a particular learner, and whether a collection covers distinct rather than repetitive situations and behaviors.

The resulting literature remains fragmented and difficult to compare. Studies commonly describe data in domain-specific forms, such as API calls~\citep{qinToolLLMFacilitatingLarge2023,liuAPIGenAutomatedPipeline2024}, repository tasks~\citep{pan2024training,jain2025r2e}, GUI demonstrations~\citep{sunOSGenesis2024,qinUITARS2025}, simulator rollouts~\citep{wangRoboGenTowardsUnleashing2023,wangGenSimGeneratingRobotic2023}, or scientific discovery experience~\citep{xuSciDisco2026,liAutoSDT2025}, and organize methods by application domain, training stage, or environment type. This organization creates two structural difficulties. First, it conflates mechanism with application: related mechanisms may be discussed separately across diverse scenarios. Second, descriptions of a pipeline often mix how candidates are constructed with how they are verified, selected, or allocated to a learner.
% , even though these operations address different questions.
As a result, similar mechanisms appear under different terminology, while data produced through substantially different processes are frequently reported under heterogeneous and difficult-to-align criteria. The agentic data generation field therefore needs a common account of both \emph{what components a pipeline constructs} and \emph{how the resulting distribution is shaped under ACE}.

To make ACE comparable across domains, we introduce a compact but sufficiently expressive factorization of agentic data as $d=(E,q,\tau,v)$. Here, $E$ describes the actionable world, including its state, dynamics, policies, and action--observation interfaces; $q$ specifies the task-conditioned objective and constraints; $\tau$ records a realized interaction; and the optional $v$ supplies outcome- or process-level supervision, including executable checks, model judgments, or rewards. This abstraction accommodates diverse agentic settings without requiring every dataset to serialize the four factors in the same way. More importantly, it makes the targets of ACE explicit: accuracy concerns local validity and consistency among the factors, complexity concerns the burden induced by their configuration, and diversity concerns their non-redundant coverage.

\begin{figure}[t]
    \centering
    \includegraphics[width=1\linewidth]{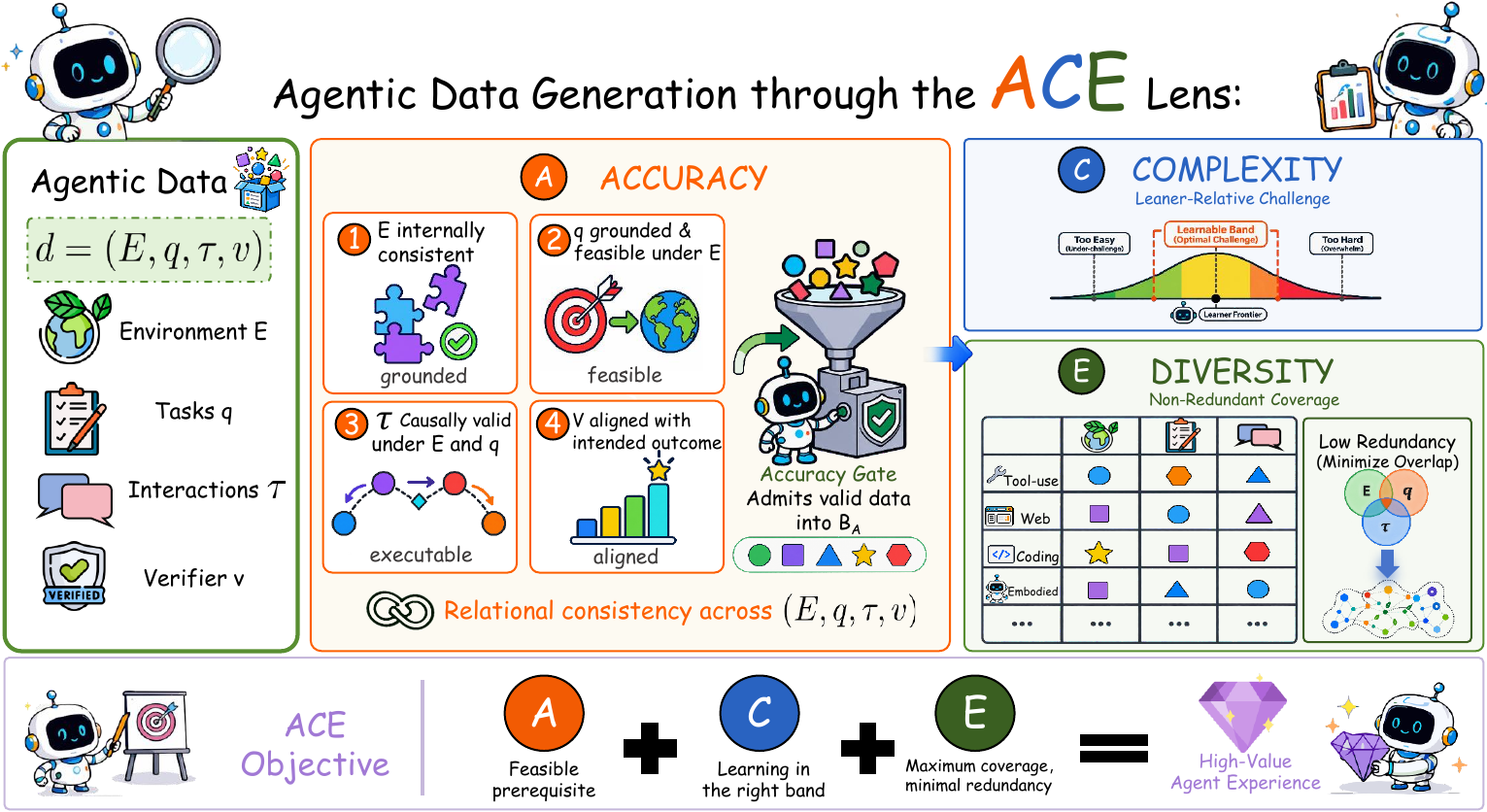}
    \vspace{-0.5cm}
    \caption{The ACE lens views agentic data generation as constrained distribution design: Accuracy defines valid support, while learner-relative Complexity and Diversity shape useful, non-redundant experience over $(E,q,\tau,v)$.}
    \label{fig:ace-lens}
\end{figure}
As shown in Figure~\ref{fig:ace-lens}, we formalize ACE as a constrained distribution-design objective. Accuracy defines the feasible support by requiring local validity and relational consistency among $E$, $q$, $\tau$, and $v$. Within that support, complexity shapes where probability mass is placed relative to the capability of a declared learner and execution configuration, while diversity determines how broadly it covers distinct, non-redundant environments, tasks, and interaction behaviors. This asymmetry is essential: difficulty or variation cannot compensate for invalidity as a generation objective, whereas uniformly valid but trivial or repetitive data may provide little additional learning value. 
% It also explains why common proxies are incomplete: final-answer correctness does not establish interaction accuracy, sequence length does not establish learner-relative complexity, and dataset size or linguistic variation does not establish behavioral diversity.
It is worth noting that ACE is not intended as an exhaustive checklist for dataset governance. Cost, efficiency, safety, etc., remain important constraints. Rather, ACE captures three generation-time properties that most directly determine whether the agentic experience is useful for learning. It consequently serves as an analytical lens over diverse generation mechanisms.
% rather than a partition of papers into disjoint method families: the same verified blueprint, executable environment, or feedback loop may affect multiple ACE dimensions.

The factorization also supports a mechanism-oriented taxonomy of generation paradigms. We organize pipelines by the factor that serves as the primary anchor and by the dependency structure through which the remaining factors are instantiated. Forward pipelines ground tasks and interactions in an existing environment; task-first and trajectory-first pipelines construct the remaining factors around a target capability or observed experience; and structure-first pipelines use an intermediate graph, plan, or blueprint to coordinate them. This view accommodates joint, iterative, and self-improving systems in which verification failures or learner feedback revise previously generated factors. It therefore describes how data are constructed, while ACE specifies which generated data should be admitted or emphasized.

Viewed through this lens, the literature reveals a coherent shift from plausibility judgments toward execution-grounded accuracy, from static difficulty heuristics toward model- and configuration-relative complexity, and from surface variation toward behavioral coverage~\citep{liuAPIGenAutomatedPipeline2024,xuEnvFactoryScalingToolUse2026,guoGenEnv2025,chenAgentFrontier2025,chenDIVEScalingDiversity2026}. Meanwhile, agentic data are expanding beyond fixed post-training trajectories toward pre- and mid-training supervision and closed-loop experience that changes with the learner~\citep{zengDaVinciDev2026,zuoQwenAgentWorldLanguageWorld2026,zhaiAgentEvolverEfficientSelfEvolving2025}. These developments make it increasingly important to separate the architecture of data generation from the objectives used to evaluate and allocate its outputs.

Our main contributions are as follows:

% \noindent\emph{1. An ACE-centered synthesis of agentic data generation.} We organize prior work around how generation pipelines ensure Accuracy, construct and calibrate learner-relative Complexity, and expand effective Diversity, together with domain-specific evidence, measurement choices, costs, and limitations.

% \noindent\emph{2. A cross-domain formulation of agentic data.} We define a common factorized data object $(E,q,\tau,v)$ that makes the generated components and their consistency relations explicit across tool use, software engineering, GUI, embodied, social, and scientific agents.

% \noindent\emph{3. A mechanism-oriented taxonomy of generation paradigms.} We organize pipelines by their primary anchor and dependency structure rather than by domain labels, while separating how data are constructed from how generated candidates are assessed and selected.

% \noindent\emph{4. A discussion of emerging data regimes.} We examine the complementary roles of real and synthetic data, the extension of agentic supervision to pre-training and mid-training, and the changing role of data generation in self-evolving agents.
\begin{tcolorbox}[
    enhanced,
    title={Main Contributions},
    colback=myblue!4,
    colbacktitle=myblue!92!black,
    colframe=myblue!88!black,
    coltitle=white,
    fonttitle=\bfseries,
    boxrule=1.1pt,
    arc=3mm,
    outer arc=3mm,
    left=7pt,
    right=7pt,
    top=7pt,
    bottom=7pt,
    toptitle=4pt,
    bottomtitle=4pt,
    lefttitle=7pt,
    righttitle=7pt,
    titlerule=0pt,
    label={box:main-contributions}
]

\noindent\textbf{1. An ACE-centered synthesis of agentic data generation.}
We organize prior work around how generation pipelines ensure Accuracy,
construct and calibrate learner-relative Complexity, and expand effective
Diversity, together with domain-specific evidence, measurement choices,
costs, and limitations.

\medskip

\noindent\textbf{2. A cross-domain formulation of agentic data.}
We define a common factorized data object $(E,q,\tau,v)$ that makes the
generated components and their consistency relations explicit across tool
use, software engineering, GUI, embodied, social, and scientific agents.

\medskip

\noindent\textbf{3. A mechanism-oriented taxonomy of generation paradigms.}
We organize pipelines by their primary anchor and dependency structure
rather than by domain labels, while separating how data are constructed
from how generated candidates are assessed and selected.

% \medskip

% \noindent\textbf{4. A discussion of emerging data regimes.}
% We discuss ACE-aware scaling, the complementary roles of real and synthetic data, the extension
% of agentic supervision to pre-training and mid-training, and the changing
% role of data generation in self-evolving agents.

\end{tcolorbox}
The remainder of this paper is organized as follows. Section~\ref{sec:formulation} defines agentic interaction and the common data object. Section~\ref{sec:gen-paradigm} introduces factorized generation paradigms and the ACE objective. Sections~\ref{sec:accuracy}--\ref{sec:diversity} synthesize existing methods and findings along the three ACE dimensions. Section~\ref{sec:discussion} then discuss scaling under ACE objective, data provenance, earlier training stages, and continual self-evolution beyond the task-level setting.

\section{Formulation}
\label{sec:formulation}

% This section establishes a common object for comparing agentic data generation methods. The purpose is not to prescribe one implementation, but to identify the parts that a generation pipeline must construct, connect, and validate.

\subsection{Agentic Multi-turn Interaction}
\label{sec:agentic-interaction}

Following prior work that models tool-augmented interaction as a partially observable Markov decision process (POMDP)~\citep{prabhakarAPIGenMTAgenticPipeline2025,zengToolACEMTNonAutoregressiveGeneration2026,dongAgentWorldScalingRealWorld2026}, we write
\begin{equation}
    \mathcal{M}=(\mathcal{U},\mathcal{S},\mathcal{A},\mathcal{O},\mathcal{P},\mathcal{R}),
    \label{eq:agentic-pomdp}
\end{equation}
where $\mathcal{U}$ is the task or user-intent space, $\mathcal{S}$ is the latent state space, $\mathcal{A}$ denotes action space which contains language responses and environment-facing actions, $\mathcal{O}$ is the observation space, $\mathcal{P}$ specifies state transitions, and $\mathcal{R}$ is an optional reward or evaluation function. At turn $t$, an agent chooses an action from the observable history, and the environment returns a new observation while possibly changing its latent state.

More concretely, the policy acts on the observable history rather than the latent state, and the selected action induces the next state and observation:
\begin{equation}
    h_t=(o_0,a_1,o_1,\ldots,a_{t-1},o_{t-1}),\qquad
    a_t\sim\pi_\theta(\cdot\mid h_t),\qquad
    (s_{t+1},o_{t+1})\sim\mathcal{P}(\cdot\mid s_t,a_t).
    \label{eq:agentic-policy-transition}
\end{equation}
This distinction is central to agentic data: the trajectory records what the policy observes and does, whereas an executable environment may additionally maintain latent state for transition and reward computation.

This abstraction covers nearly all agent domains. In tool use, actions are structured API calls and observations are tool returns. In web and GUI agents, they are interface operations and visual or structured page states. In coding agents, the environment includes a repository, shell, dependencies, and tests, while in embodied settings, it may be a simulator or physical world. User turns are included in observations: they progressively reveal constraints, answer clarification questions, or revise an earlier request. Consequently, the task should not always be identified with the first utterance; it can be a latent intent expressed over the interaction.

A realized interaction is denoted by
\begin{equation}
    \tau=(o_0,a_1,o_1,\ldots,a_T,o_T),
    \label{eq:interaction-trajectory}
\end{equation}
where $a_i$ are actions including text responses and tool calls, observations $o_j$ can be user messages or environment feedback. This representation accommodates both simple function-calling examples and long, stateful, multi-turn trajectories.

\subsection{Environment Parameterization}
\label{sec:environment-parameterization}

% The POMDP abstracts away how an actionable world is implemented. 
For data generation, a concrete environment can be parameterized as
\begin{equation}
    e=(\mathcal{D},\mathcal{F},\mathcal{P}_{\mathrm{rule}},\Omega,v),
    \label{eq:environment-parameterization}
\end{equation}
where $\mathcal{D}$ is an optional state carrier such as a database, repository, application, or simulator; $\mathcal{F}$ is the available tool or action set; $\mathcal{P}_{\mathrm{rule}}$ contains policies, permissions, and domain constraints; $\Omega$ determines which parts of the latent state are exposed as observations; and $v$ is an optional success interface. A serialized environment specification $E$ describes some or all of these components. In simple function calling, $E$ may contain only tool schemas and usage rules; in RL-oriented data, $E$ must additionally support outcome evaluation.

\subsection{From Interaction to Agentic Data}
\label{sec:from-interaction-to-data}

Agentic data instantiates the interaction process through three core factors.

\paragraph{Environment Specification ($E$).}
The environment describes the actionable world. It may contain tool schemas, a database or simulator state, transition rules, policies, permissions, observation interfaces, and termination conditions. Some datasets serialize only textual tool descriptions, whereas executable environments expose state transitions and can support new rollouts. Thus, $E$ ranges from a static interface specification to a complete interaction substrate.

\paragraph{Task Signal ($q$).}
The task specifies what should be achieved and under which constraints. It may be an explicit instruction, a target state, a hidden user intent, or a signal progressively expressed across several turns. A task is grounded only relative to an environment: its entities must exist, required actions must be available, and its completion condition must be meaningful under the environment rules.

\paragraph{Interaction Realization ($\tau$).}
The realization specifies how the task is attempted or solved in the environment. For supervised fine-tuning, it is commonly a demonstrated trajectory. For reinforcement learning, it may instead be sampled online from an executable environment. A pipeline can also retain multiple successful, failed, exploratory, or recovery trajectories for the same environment--task pair.

These factors give the central decomposition used in this survey:
% \begin{equation}
%     \text{Agentic Data}=\text{Environment Specification}
%     +\text{Task Signal}+\text{Interaction Realization}.
%     \label{eq:agentic-data-components}
% \end{equation}

\begin{tcolorbox}[
    enhanced,
    title={Agentic Data Decomposition},
    colback=myblue!4,
    colbacktitle=myblue!92!black,
    colframe=myblue!88!black,
    coltitle=white,
    fonttitle=\bfseries,
    boxrule=1.1pt,
    arc=3mm,
    outer arc=3mm,
    left=7pt,
    right=7pt,
    top=7pt,
    bottom=7pt,
    toptitle=4pt,
    bottomtitle=4pt,
    lefttitle=7pt,
    righttitle=7pt,
    titlerule=0pt,
    label={box:agent-data}
]
\begin{equation}
    \text{Agentic Data}
    =
    \text{Environment Specification}
    +
    \text{Task Signal}
    +
    \text{Interaction Realization}.
    \label{eq:agentic-data-components}
\end{equation}
\end{tcolorbox}
The decomposition is conceptual rather than a serialization requirement. For example, the initial state may be stored inside an executable environment, and a progressively revealed task may be represented through the user turns in $\tau$.

\subsection{A Common Data Object}
\label{sec:agentic-data-object}
\begin{tcolorbox}[
    enhanced,
    title={Common Data Object},
    colback=myblue!4,
    colbacktitle=myblue!92!black,
    colframe=myblue!88!black,
    coltitle=white,
    fonttitle=\bfseries,
    boxrule=1.1pt,
    arc=3mm,
    outer arc=3mm,
    left=7pt,
    right=7pt,
    top=7pt,
    bottom=7pt,
    toptitle=4pt,
    bottomtitle=4pt,
    lefttitle=7pt,
    righttitle=7pt,
    titlerule=0pt
]
For the remainder of the survey, we use the common data object
\begin{equation}
    d=(E,q,\tau,v),
    \label{eq:agentic-data-object}
\end{equation}
where $v$ is an optional verifier or reward interface. The first three factors define the interactive problem and its realization; $v$ records how their consistency or outcome is evaluated. It may be a schema checker, executable test, terminal-state predicate, policy rule, proof assistant, LLM judge, or a hybrid of these mechanisms. Treating $v$ as an interface avoids conflating the object being generated with one particular training paradigm.

\end{tcolorbox}

For SFT, $d$ usually stores $E$ and a fixed $\tau$, with $q$ expressed explicitly or through its user turns; $v$ is optional and may only be used during curation. For RL and environment-based evaluation, $E$ must support new interactions, while $q$ and $v$ define the rollout objective and success condition. The same notation therefore allows us to compare trajectory synthesis, executable environment construction, and online task generation without claiming that their released artifacts are identical.

\begin{figure}[htbp]
    \centering
    \includegraphics[width=\textwidth]{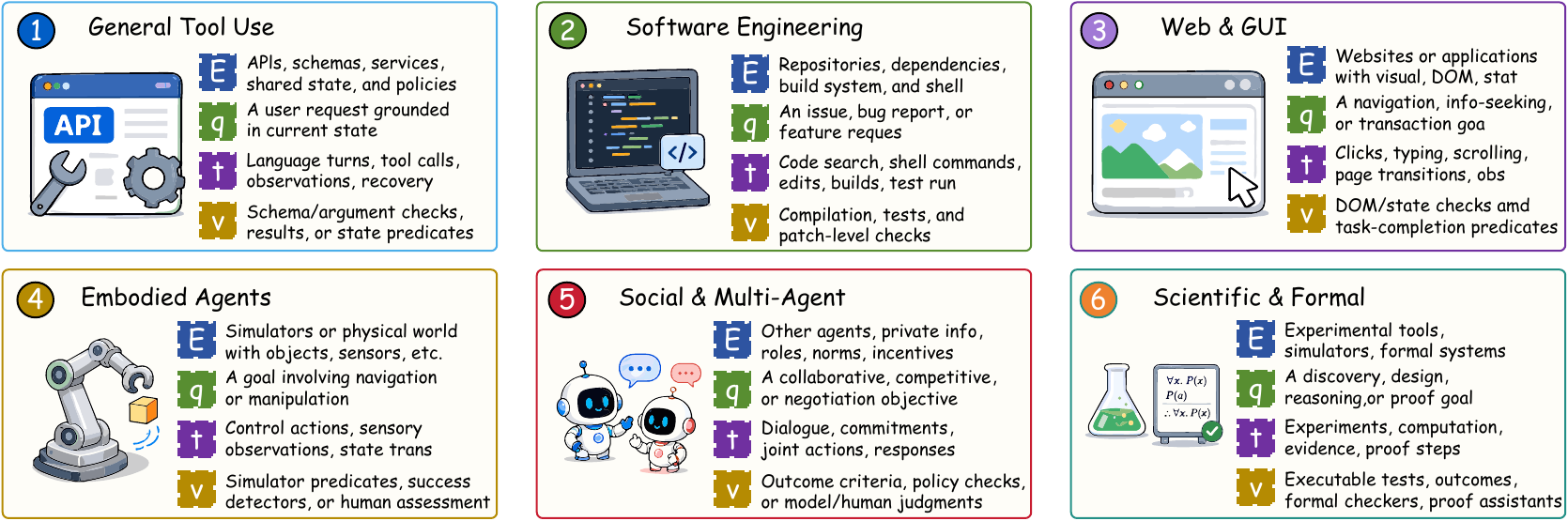}
    \caption{Cross-domain instantiations of the common agentic data object $d=(E,q,\tau,v)$. The factors retain the same conceptual roles even though their concrete representations and verification signals differ across domains.}
    \label{fig:agentic-data-across-domains}
\end{figure}

Figure~\ref{fig:agentic-data-across-domains} illustrates how the same factors are instantiated across representative agent domains. Their concrete forms differ, but each domain requires an actionable world, a grounded objective, a realized or sampled interaction, and some basis for assessing validity or success.

This factorization also clarifies the scope of agentic data generation. It is not simply instruction generation followed by response sampling. The pipeline must establish compatible relations among an actionable environment, a grounded task, an interaction process, and, when needed, a trustworthy success signal. 
% Sections~\ref{sec:accuracy}--\ref{sec:diversity} examine how existing work controls the accuracy, complexity, and diversity of these coupled factors.

\section{Generation Paradigms}
\label{sec:gen-paradigm}

Section~\ref{sec:formulation} identifies the factors contained in agentic data. We next organize existing generation pipelines by the order in which they construct and ground these factors. 
% This taxonomy describes the pipeline architecture; it does not assign each work exclusively to one ACE dimension. The same pipeline may contain mechanisms that improve accuracy, calibrate complexity, and expand diversity simultaneously.

\subsection{A Factorized View}
\label{sec:factorized-view}

Agentic data generation can be viewed as designing a joint distribution over environments, tasks, and interaction realizations. The dominant factorization is
\begin{equation}
    p(E,q,\tau)=p(E)\,p(q\mid E)\,p(\tau\mid E,q),
    \label{eq:forward-factorization}
\end{equation}
but this order is not mandatory. Task-first pipelines begin from $q$ and construct a compatible $E$; trajectory-first pipelines recover $q$ from an executable or observed $\tau$.
% structure-first pipelines introduce a graph, plan, or blueprint before realizing the final interaction. 
The ordering determines which factors are directly controlled and where consistency must be recovered.

\begin{figure}[t]
    \centering
    \includegraphics[
        page=1,
        width=\textwidth,
        keepaspectratio
    ]{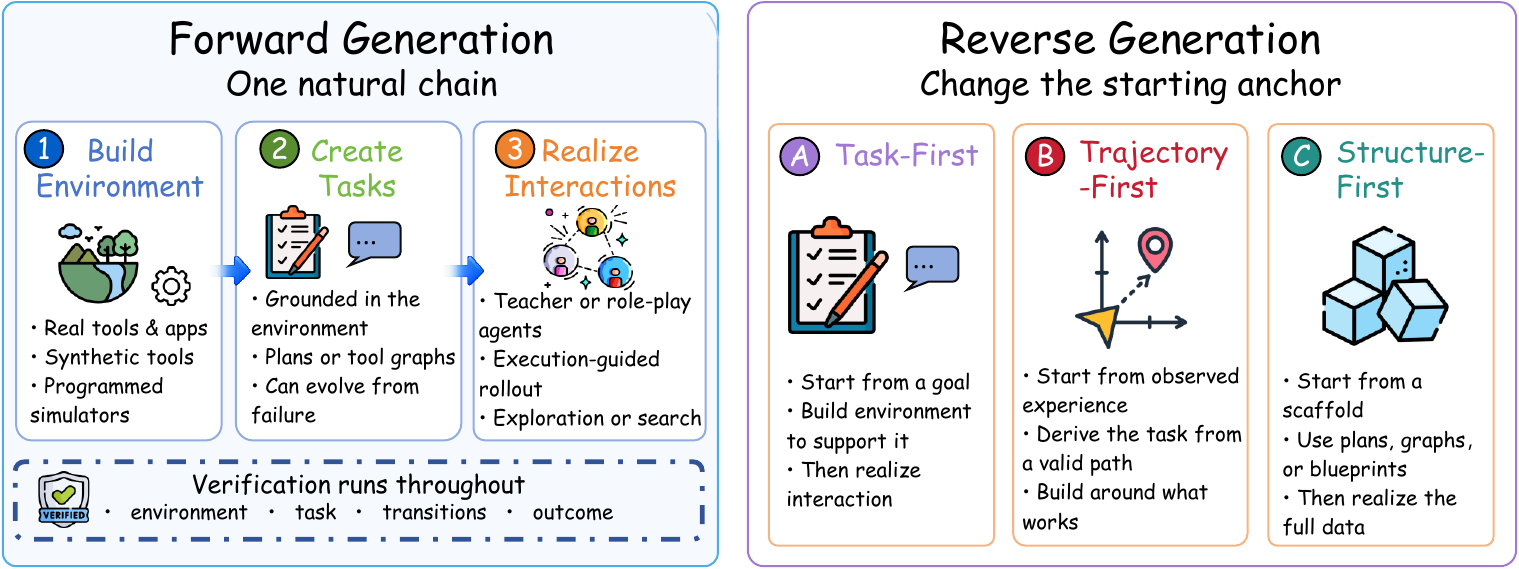}
    \caption{Generation Paradigms: Forward Generation vs Reverse Generation.}
    \label{fig:generation-paradigms}
\end{figure}

This view separates two questions that are often conflated. A \emph{generation paradigm} specifies how the factors are produced. A \emph{data objective} specifies which generated instances should be accepted or emphasized. The former is discussed below, which can be categoried into twofolds, \emph{forward generation} and \emph{reverse generation} (see Figure~\ref{fig:generation-paradigms}); the ACE objective is introduced at the end of this section serving as principles during generation.

\subsection{Forward Generation}
\label{sec:forward-generation}

Forward generation constructs an environment, generates grounded tasks based on the environments, and then obtains trajectories. It mirrors the natural dependency of interaction: available actions and states determine which tasks are feasible, and both determine which trajectories are meaningful.

\paragraph{Constructing Environments.}
Existing work obtains $E$ from three overlapping sources. First, real specifications are crawled or curated from existing APIs, MCP services, repositories, websites, and applications. A substantial body of prior work grounds data generation in real-world tool ecosystems~\citep{qinToolLLMFacilitatingLarge2023,patilGorillaLargeLanguage2023,liuAPIGenAutomatedPipeline2024,shimToolDialMultiturnDialogue2025,liCloseLoopSynthesizing2025,xuTOUCANSynthesizing15M2025}, and similarly, software-engineering pipelines instantiate execution environments from repositories collected from real-world projects~\citep{pan2024training,jain2025r2e,yang2026swe,badertdinov2026swe}. While these sources offer high realism, they also inherit uneven documentation, unstable dependencies, and constraints related to execution, access, and licensing.

Second, LLM synthesis expands tools, rules, and domain descriptions beyond collected resources. ToolACE~\citep{liuToolACEWinningPoints2025} evolves an API pool before generating tasks and trajectories, while ToolAlpaca~\citep{tangToolAlpacaGeneralizedTool2023}, Seal-Tools~\citep{wuSealToolsSelfInstructTool2024}, SynthTools~\citep{castellani2025synthtools}, and ToolWeave~\citep{khandelwal2026toolweave} use related forms of tool or tool-graph synthesis. This route is scalable and controllable, but generated interfaces can be underspecified or detached from plausible workflows.

Third, programmatic construction explicitly implements databases, transition dynamics, simulators, and validators. This line of work marks a shift from static textual tool descriptions toward stateful, resettable, and verifiable environments~\citep{songEnvScalerScalingToolInteractive2026,wangAgentWorldModel2026,dongAgentWorldScalingRealWorld2026,xuEnvFactoryScalingToolUse2026,tuScaleEnvScalingEnvironment2026,tianASTRAAutomatedSynthesis2026,zhouToolVerse2026}. The same paradigm has also been extended to executable tool-use arenas, scientific-discovery environments, and simulation-ready 3D worlds~\citep{duCodeGym2025,xuSciDisco2026,wangEmbodiedGenV22026,zhouEmbodiedClaw2026}. 
Programmatic construction offers stronger support for stateful interaction, controlled reset, and reliable verification. However, this flexibility comes at a higher engineering cost. Environments must be implemented, tested, and maintained, while the breadth and fidelity of the resulting domains depend heavily on the quality of the underlying code generation or environment design.
% Programmatic environments support both offline trajectory collection and online RL rollouts, at the cost of implementation, testing, and maintenance. In practice, many systems combine real specifications, LLM expansion, executable reconstruction, and environments derived from existing repositories or datasets.

\paragraph{Generating Grounded Tasks.}
The simplest approach prompts a model to write instructions from tool descriptions~\citep{qinToolLLMFacilitatingLarge2023,liuAPIGenAutomatedPipeline2024}. More recent work introduces intermediate grounding structures. Tool-graph-guided (or skill graph) methods sample compatible tools or dependency paths before writing the request~\citep{wangToolFlowBoostingLLM2025,shimToolDialMultiturnDialogue2025,yinMagnetMultiturnTooluse2025,yangToolMindTechnicalReport2025,xuEnvFactoryScalingToolUse2026,fanSkillSynth2026}. Blueprint- and plan-first methods specify intended actions or subgoals before dialogue realization~\citep{prabhakarAPIGenMTAgenticPipeline2025,wangToolFlowBoostingLLM2025,erdoganPlanAndAct2025}. Stateful systems jointly construct the initial state, task, and verifier so that feasibility is established in a particular environment instance~\citep{caiAutoForgeAutomatedEnvironment2025,songEnvScalerScalingToolInteractive2026,tuScaleEnvScalingEnvironment2026,dongAgentWorldScalingRealWorld2026}. Other pipelines evolve existing tasks or derive new ones from exploration and failure traces, making generation more targeted to current capability gaps~\citep{huAgentGenEnhancingPlanning2025,zhaiAgentEvolverEfficientSelfEvolving2025,haoFailureMasteryGenerating2026,fuSESA2026,xiaoSocraticSWE2026}.

\paragraph{Realizing Interactions.}
Trajectories can be produced by a teacher model, role-playing agents, execution-guided rollout, or combinations of them. Earlier function-calling datasets often realize short interactions from an explicit task~\citep{qinToolLLMFacilitatingLarge2023,patilGorillaLargeLanguage2023,liuAPIGenAutomatedPipeline2024}. Multi-turn pipelines simulate user, assistant, and sometimes tool roles to expose constraints progressively and collect clarification or recovery behavior~\citep{tangToolAlpacaGeneralizedTool2023,liuToolACEWinningPoints2025,wangToolFlowBoostingLLM2025,prabhakarAPIGenMTAgenticPipeline2025,liCloseLoopSynthesizing2025,guWRITWriteRead2026}. Executable environments instead return observations from explicit state transitions and can reject invalid actions during rollout~\citep{songEnvScalerScalingToolInteractive2026,xuEnvFactoryScalingToolUse2026,tianASTRAAutomatedSynthesis2026,duCodeGym2025}. Exploration-driven web and mobile pipelines further collect multimodal trajectories and recovery behavior from interaction rather than relying entirely on scripted demonstrations~\citep{pahujaExplorer2025,chengOpenMobile2026,qinUITARS2025}. Search and deep-research agents collect another form of interaction realization in which retrieval, evidence integration, and repeated query reformulation unfold under online or real-world feedback~\citep{zhengDeepResearcher2025,liWebSailorV22025,jinSearchR12025,songR1Searcher2025}. Structural scaffolds such as plans and tool graphs improve long-range coherence before local utterances are generated.

When environment implementation is the bottleneck, LLM-based simulators can generate stateful responses from API specifications and interaction history~\citep{liSimia2025,leeEnvironmentFree2026}, and even serve as rehearsal environments during learning~\citep{xuEnvACE2026}.

Table~\ref{tab:forward-generation} summarizes the representative works. This kind of paradigm offers strong grounding because later factors are conditioned on an existing environment. Its main risk is cascading dependence: narrow or unreliable environments constrain every task and trajectory built on top of them. Verification is therefore not merely a final step. Recent pipelines increasingly test environment components, task feasibility, intermediate transitions, and terminal outcomes throughout generation. Section~\ref{sec:accuracy} further surveys these mechanisms.
\definecolor{groupblue}{RGB}{214,232,248}
\definecolor{githubpink}{RGB}{230,45,110}

% ============================================================
% GitHub link
% ============================================================

\newcommand{\githubrepo}[1]{%
  \href{#1}{%
    \textcolor{black}{\faGithub}\,
    \textcolor{githubpink}{GitHub}%
  }%
}

\newcommand{\huggingfacerepo}[1]{%
  \href{#1}{%
    \raisebox{-0.3em}{\includegraphics[height=1.1em]{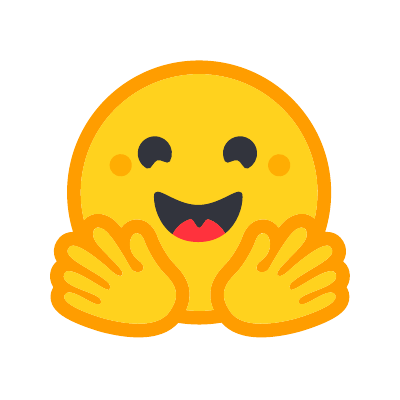}}\,
    \textcolor{githubpink}{Hugging Face}%
  }%
}

% ============================================================
% Float settings
% ============================================================

\renewcommand{\topfraction}{0.95}
\renewcommand{\dbltopfraction}{0.95}
\renewcommand{\textfraction}{0.05}
\renewcommand{\floatpagefraction}{0.90}
\renewcommand{\dblfloatpagefraction}{0.90}

% ============================================================
% Forward Generation Table
% ============================================================

% ============================================================
% Forward Generation Table
% ============================================================

% ============================================================
% Forward Generation Table
% ============================================================

\begin{table*}[!t]
\centering

\caption{
\textbf{Representative works in forward agentic data generation.}
All methods follow the dependency $E \rightarrow q \rightarrow \tau$.
Methods are grouped according to the dominant source of environment
construction: real or curated environments, LLM-synthesized environments,
and programmatic or executable environments.
\textbf{Resource} links to public GitHub repositories or Hugging Face websites when available.
}
\label{tab:forward-generation}

% ------------------------------------------------------------
% Overall table style
% ------------------------------------------------------------

\footnotesize

% Horizontal padding between columns
\setlength{\tabcolsep}{3.2pt}

% Vertical spacing between rows
\renewcommand{\arraystretch}{1.08}

% ============================================================
% Adaptive-width table
%
% Relative widths:
% Method            1.20
% E Construction    1.05
% q Generation      1.05
% tau Realization   1.15
% Resource          0.55
%
% Sum = 5.00
% ============================================================

\begin{tabularx}{\linewidth}{@{}
>{\hsize=1.20\hsize\linewidth=\hsize
  \raggedright\arraybackslash}X
>{\hsize=1.05\hsize\linewidth=\hsize
  \raggedright\arraybackslash}X
>{\hsize=1.05\hsize\linewidth=\hsize
  \raggedright\arraybackslash}X
>{\hsize=1.15\hsize\linewidth=\hsize
  \raggedright\arraybackslash}X
>{\hsize=0.55\hsize\linewidth=\hsize
  \centering\arraybackslash}X
@{}}

\toprule

\textbf{Method}
&
\textbf{$E$ Construction}
&
\textbf{$q$ Generation}
&
\textbf{$\tau$ Realization}
&
\textbf{Resource}
\\

\midrule

% ============================================================
% Real / Curated Environments
% ============================================================

\rowcolor{groupblue}
\multicolumn{5}{@{}l@{}}{%
  \rule{0pt}{2.15ex}\textbf{Real / Curated Environments}
}
\\

ToolLLM~\citep{qinToolLLMFacilitatingLarge2023}
&
Real API collection
&
Tool-description prompting
&
Teacher-guided tool rollout
&
\githubrepo{https://github.com/OpenBMB/ToolBench}
\\

Gorilla~\citep{patilGorillaLargeLanguage2023}
&
Real API collection
&
API-grounded instructions
&
Single-turn API invocation
&
\githubrepo{https://github.com/ShishirPatil/gorilla}
\\

APIGen~\citep{liuAPIGenAutomatedPipeline2024}
&
Executable API pool
&
Verified task synthesis
&
Executable function-call rollout
&
\githubrepo{https://github.com/SalesforceAIResearch/xLAM}
\\

ToolDial~\citep{shimToolDialMultiturnDialogue2025}
&
Real API graph
&
Graph-guided task synthesis
&
Multi-turn role-playing
&
\githubrepo{https://github.com/holi-lab/ToolDial}
\\

InfTool~\citep{liCloseLoopSynthesizing2025}
&
Connected executable APIs
&
Tool-conditioned synthesis
&
Multi-turn role-playing
&
--
\\

TOUCAN~\citep{xuTOUCANSynthesizing15M2025}
&
Real MCP environments
&
MCP-grounded task synthesis
&
Real tool execution rollout
&
\githubrepo{https://github.com/TheAgentArk/Toucan}
\\

% ============================================================
% LLM-Synthesized Environments
% ============================================================

\addlinespace[4pt]
\midrule

\rowcolor{groupblue}
\multicolumn{5}{@{}l@{}}{%
  \rule{0pt}{2.15ex}\textbf{LLM-Synthesized Environments}
}
\\

ToolAlpaca~\citep{tangToolAlpacaGeneralizedTool2023}
&
LLM-synthesized tools
&
Tool-conditioned instructions
&
Multi-agent simulation
&
\githubrepo{https://github.com/tangqiaoyu/ToolAlpaca}
\\

ToolACE~\citep{liuToolACEWinningPoints2025}
&
Self-evolved API pool
&
Tool-grounded task synthesis
&
Multi-agent dialogue rollout
&
\huggingfacerepo{https://huggingface.co/datasets/Team-ACE/ToolACE}
\\

Seal-Tools~\citep{wuSealToolsSelfInstructTool2024}
&
Synthetic tool generation
&
Self-instruct task synthesis
&
Simulated tool-use rollout
&
\githubrepo{https://github.com/fairyshine/Seal-Tools}
\\

SynthTools~\citep{castellani2025synthtools}
&
Hierarchical environment synthesis
&
Verifiable task synthesis
&
Simulator-grounded rollout
&
\githubrepo{https://github.com/namkoong-lab/SynthTools}
\\

ToolWeave~\citep{khandelwal2026toolweave}
&
Synthetic tool graphs
&
Graph-guided goal synthesis
&
Plan-guided dialogue synthesis
&
\githubrepo{https://github.com/IBM/ToolWeave}
\\

% ============================================================
% Programmatic / Executable Environments
% ============================================================

\addlinespace[4pt]
\midrule

\rowcolor{groupblue}
\multicolumn{5}{@{}l@{}}{%
  \rule{0pt}{2.15ex}\textbf{Programmatic / Executable Environments}
}
\\

AutoForge~\citep{caiAutoForgeAutomatedEnvironment2025}
&
Programmatic environment synthesis
&
State-conditioned task synthesis
&
Executable environment rollout
&
--
\\

EnvScaler~\citep{songEnvScalerScalingToolInteractive2026}
&
Programmatic tool environments
&
State-conditioned scenarios
&
Executable agent rollout
&
\githubrepo{https://github.com/RUC-NLPIR/EnvScaler}
\\

% Agent World Model~\citep{wangAgentWorldModel2026}
% &
% Database-backed environment synthesis
% &
% State-grounded task synthesis
% &
% MCP-based agent rollout
% &
% \githubrepo{https://github.com/Snowflake-Labs/agent-world-model}

Agent-World~\citep{dongAgentWorldScalingRealWorld2026}
&
Executable world construction
&
State-grounded task synthesis
&
Online agent rollout
&
--
\\

EnvFactory~\citep{xuEnvFactoryScalingToolUse2026}
&
Executable environment factory
&
Graph- and state-grounded tasks
&
Executable multi-turn rollout
&
\githubrepo{https://github.com/LARK-AI-Lab/EnvFactory}
\\

ScaleEnv~\citep{tuScaleEnvScalingEnvironment2026}
&
Programmatic environment scaling
&
State-conditioned task synthesis
&
Executable RL rollout
&
--
\\

\bottomrule

\end{tabularx}

\end{table*}

\definecolor{groupblue}{RGB}{214,232,248}
\definecolor{githubpink}{RGB}{230,45,110}

\definecolor{groupblue}{RGB}{214,232,248}
\definecolor{githubpink}{RGB}{230,45,110}

\newcolumntype{M}{%
  >{\hsize=1.45\hsize\linewidth=\hsize
    \raggedright\arraybackslash}X%
}

\newcolumntype{D}{%
  >{\hsize=0.55\hsize\linewidth=\hsize
    \raggedright\arraybackslash}X%
}

\newcolumntype{K}{%
  >{\hsize=1.45\hsize\linewidth=\hsize
    \raggedright\arraybackslash}X%
}

\newcolumntype{R}{%
  >{\hsize=0.55\hsize\linewidth=\hsize
    \centering\arraybackslash}X%
}

% ============================================================
% Table
% ============================================================

\begin{table*}[!t]
\centering

\caption{
\textbf{Representative works in reverse agentic data generation.}
Methods are grouped into task-first, trajectory-first, and structure-first
generation, followed by adaptive and self-evolving approaches as a
cross-cutting extension. \textbf{Domain} indicates the primary application
setting, while \textbf{Resource} links to public repositories when available.
}
\label{tab:reverse-generation}

% ------------------------------------------------------------
% Overall table style
% ------------------------------------------------------------
\footnotesize

% Horizontal padding between columns
\setlength{\tabcolsep}{4pt}

% Vertical spacing between rows
\renewcommand{\arraystretch}{1.12}

% ============================================================
% Adaptive table: automatically follows current page width
% ============================================================
\begin{tabularx}{\linewidth}{@{} M D K R @{}}

\toprule

\textbf{Method}
&
\textbf{Domain}
&
\textbf{Key Mechanism}
&
\textbf{Resource}
\\

\midrule

% ============================================================
% Task-First Generation
% ============================================================

\rowcolor{groupblue}
\multicolumn{4}{@{}l@{}}{%
  \rule{0pt}{2.15ex}\textbf{Task-First Generation}
}
\\

AgentInstruct~\citep{mitraAgentInstructGenerativeTeaching2024}
&
General
&
Capability-targeted environment synthesis
&
--
\\

Agentic Proposing~\citep{jiaoAgenticProposingEnhancing2026}
&
General
&
Problem-first downstream synthesis
&
\githubrepo{https://github.com/Frostlinx/Agentic-Proposing}
\\

BUTTON~\citep{chenFacilitatingMultiturnFunction2025}
&
Tool
&
Compositional multi-turn task synthesis
&
\githubrepo{https://github.com/PKU-Baichuan-MLSystemLab/BUTTON}
\\

ToolBridge~\citep{jin2024toolbridge}
&
Tool
&
Existing-task tool integration
&
\githubrepo{https://github.com/CharlesPikachu/ToolBridge}
\\

ToRA~\citep{gou2024tora}
&
Math
&
Tool-integrated mathematical reasoning
&
\githubrepo{https://github.com/microsoft/ToRA}
\\

MathCoder~\citep{wang2024mathcoder}
&
Math
&
Code-assisted mathematical reasoning
&
\githubrepo{https://github.com/mathllm/MathCoder}
\\

MARIO~\citep{liao2024mario}
&
Math
&
Code-interpreter augmented reasoning
&
\githubrepo{https://github.com/MARIO-Math-Reasoning/MARIO}
\\

AgentMath~\citep{luoAgentMath2025}
&
Math
&
Tool-augmented mathematical reasoning
&
--
\\

ReTool~\citep{fengReTool2025}
&
Math
&
Strategic tool-use reasoning
&
\githubrepo{https://github.com/ReTool-RL/ReTool}
\\

ToRL~\citep{liToRL2025}
&
Math
&
Tool-integrated reinforcement learning
&
\githubrepo{https://github.com/GAIR-NLP/ToRL}
\\

AutoSDT~\citep{liAutoSDT2025}
&
Science
&
Scientific discovery task transformation
&
\githubrepo{https://github.com/OSU-NLP-Group/AutoSDT}
\\

Agentic-Ideation~\citep{zhaoAgenticIdeation2026}
&
Science
&
Scientific reasoning-path reconstruction
&
--
\\

% ============================================================
% Trajectory-First Generation
% ============================================================

\addlinespace[4pt]
\midrule

\rowcolor{groupblue}
\multicolumn{4}{@{}l@{}}{%
  \rule{0pt}{2.15ex}\textbf{Trajectory-First Generation}
}
\\

Learn-by-interact~\citep{suLearninteractDataCentricFramework2025}
&
Tool
&
Interaction-derived task generation
&
--
\\

Trajectory2Task~\citep{wangTrajectory2TaskTrainingRobust2026}
&
Tool
&
Executable trajectory-to-task conversion
&
--
\\

Unlocking Implicit Experience~\citep{xuUnlockingImplicitExperience2026}
&
Tool
&
Implicit workflow mining from text
&
--
\\

OS-Genesis~\citep{sunOSGenesis2024}
&
GUI
&
GUI trajectory-to-task reversal
&
\githubrepo{https://github.com/OS-Copilot/OS-Genesis}
\\

Explorer~\citep{pahujaExplorer2025}
&
Web
&
Exploration-driven web task synthesis
&
\githubrepo{https://github.com/OSU-NLP-Group/Explorer}
\\

OpenMobile~\citep{chengOpenMobile2026}
&
Mobile
&
Mobile interaction task synthesis
&
\githubrepo{https://github.com/njucckevin/OpenMobile-Code}
\\

% ============================================================
% Structure-First Generation
% ============================================================

\addlinespace[4pt]
\midrule

\rowcolor{groupblue}
\multicolumn{4}{@{}l@{}}{%
  \rule{0pt}{2.15ex}\textbf{Structure-First Generation}
}
\\

ToolACE-MT~\citep{zengToolACEMTNonAutoregressiveGeneration2026}
&
Tool
&
Coarse-to-fine interaction refinement
&
--
\\

Magnet~\citep{yinMagnetMultiturnTooluse2025}
&
Tool
&
Structured tool-graph realization
&
--
\\

ToolFlow~\citep{wangToolFlowBoostingLLM2025}
&
Tool
&
Tool-graph guided task planning
&
--
\\

APIGen-MT~\citep{prabhakarAPIGenMTAgenticPipeline2025}
&
Tool
&
Verified blueprint trajectory realization
&
\huggingfacerepo{https://huggingface.co/datasets/Salesforce/APIGen-MT-5k}
\\

Execution-First~\citep{ouajdiExecutionFirstSynthetic2026}
&
Tool
&
Execution-validated task synthesis
&
--
\\

% ============================================================
% Adaptive and Self-Evolving Generation
% ============================================================

\addlinespace[4pt]
\midrule

\rowcolor{groupblue}
\multicolumn{4}{@{}l@{}}{%
  \rule{0pt}{2.15ex}\textbf{Adaptive and Self-Evolving Generation}
}
\\

AFlow~\citep{zhangAFlowAutomatingAgentic2024}
&
General
&
Search-based agentic workflow optimization
&
\githubrepo{https://github.com/FoundationAgents/AFlow}
\\

Chain-of-Agents~\citep{liChainAgentsEndEndAgent2025}
&
General
&
Multi-agent distillation and agentic RL
&
\githubrepo{https://github.com/OPPO-PersonalAI/Agent_Foundation_Models}
\\

AgentEvolver~\citep{zhaiAgentEvolverEfficientSelfEvolving2025}
&
Tool
&
Self-questioning and experience-guided evolution
&
\githubrepo{https://github.com/modelscope/AgentEvolver}
\\

WebEvolver~\citep{fangWebEvolver2025}
&
Web
&
World-model-guided web self-improvement
&
\githubrepo{https://github.com/Tencent/SelfEvolvingAgent}
\\

SESA~\citep{fuSESA2026}
&
Search
&
Self-play task posing and skill evolution
&
\githubrepo{https://github.com/Zenghuang-Fu/SESA-Self-Evolving-Search-Agents}
\\

Socratic-SWE~\citep{xiaoSocraticSWE2026}
&
SWE
&
Trace-derived skills for adaptive task generation
&
--
\\

\bottomrule

\end{tabularx}

\end{table*}
\subsection{Reverse Generation}
\label{sec:reverse-generation}

We use \emph{reverse generation} for pipelines that do not follow the direct environment--task--trajectory order (see representative works in Table~\ref{tab:reverse-generation}). They alter which artifact anchors the others and therefore offer different forms of control.

\paragraph{Task-First Generation.}
Task-first pipelines specify a target capability, instruction pattern, or composed goal and then construct the tools, environment, and interaction needed to realize it. AgentInstruct~\citep{mitraAgentInstructGenerativeTeaching2024} and Agentic Proposing~\citep{jiaoAgenticProposingEnhancing2026} use generated capability targets or problems to drive later synthesis. BUTTON composes atomic tasks into complex multi-turn requests before synthesizing corresponding functions and trajectories~\citep{chenFacilitatingMultiturnFunction2025}. One line of methods similarly transform existing mathematical, coding, or discovery tasks into tool-integrated interactions~\citep{jin2024toolbridge,gou2024tora,wang2024mathcoder,liao2024mario,luoAgentMath2025,fengReTool2025,liToRL2025,liAutoSDT2025}. Agentic-Ideation reconstructs tool-augmented reasoning paths from reference ideas, showing that the same order applies beyond digital navigation~\citep{zhaoAgenticIdeation2026}. This order provides direct control over task content and difficulty, but requires strong checks that the constructed environment actually supports the requested behavior.

\paragraph{Trajectory-First Generation.}
Trajectory-first methods explore an environment or mine a workflow before writing the user-facing task. Learn-by-interact~\citep{suLearninteractDataCentricFramework2025} and Trajectory2Task~\citep{wangTrajectory2TaskTrainingRobust2026} derive tasks from observed valid behavior, while tutorial-mining methods~\citep{xuUnlockingImplicitExperience2026} extract implicit procedures before formulating explicit instructions. OS-Genesis~\citep{sunOSGenesis2024} reverses GUI trajectories into tasks; other similar agents use interaction to expand web or mobile trajectories~\citep{pahujaExplorer2025,chengOpenMobile2026}. Grounding the task in an existing path improves executability and supports reverse construction of ambiguous, changing, or recovery-oriented intents. The resulting distribution, however, is bounded by what the exploration policy discovers.

\paragraph{Structure-First Generation.}
Another line of methods first generate an intermediate object such as a tool graph, function path, dialogue skeleton, or task blueprint. ToolACE-MT~\citep{zengToolACEMTNonAutoregressiveGeneration2026} refines a coarse interaction non-autoregressively,  Magnet~\citep{yinMagnetMultiturnTooluse2025} realizes structured tool-call graphs as multi-turn data, and Execution-First~\citep{ouajdiExecutionFirstSynthetic2026} generates template graphs and then tool-use traces before task realization. More works likewise use plans or verified blueprints to stabilize dependencies across turns~\citep{wangToolFlowBoostingLLM2025,prabhakarAPIGenMTAgenticPipeline2025}. We do not view the scaffold as a fourth data factor. It is rather a construction device that controls relations among $E$, $q$, and $\tau$.

More broadly, agentic data generation is shifting from static, one-shot synthesis toward adaptive and self-evolving processes~\citep{zhangAFlowAutomatingAgentic2024,liChainAgentsEndEndAgent2025,zhaiAgentEvolverEfficientSelfEvolving2025,fangWebEvolver2025,fuSESA2026,xiaoSocraticSWE2026}. Rather than producing independent samples with a fixed generator, these systems use accumulated experience, verification outcomes, model behavior, and observed coverage gaps to continually revise generation strategies, construct new tasks and workflows, and target the learner's evolving needs. Data generation thus becomes a closed-loop process that co-evolves with the agent and its environment.

\subsection{ACE as Constrained Distribution Design}
\label{sec:ace-distribution-design}

The paradigms above explain how a pipeline constructs candidates. The ACE lens explains how it should shape the accepted distribution. Let $p_\phi$ denote a generation pipeline and $\mathcal{B}_A$ the subset of a generated batch that passes checks on environment consistency, task feasibility, trajectory validity, etc. Accuracy is therefore an admission condition rather than a benefit that can be compensated for by other properties.

Within this valid subset, complexity should be calibrated rather than blindly maximized. We use $C_z(d)$ for the difficulty of an instance under a declared learner and execution configuration $z$, including the model, available scaffold and tools, verifier, and inference budget. A utility $g_z(C_z(d))$ can favor hard cases or, more commonly for training, a learnable band near the current model frontier. Diversity is a batch-level property $D(\mathcal{B}_A)$ that rewards coverage and non-redundancy over environments, tasks, and interaction realizations.
% \begin{equation}
%     \max_{\phi}\;
%     \mathbb{E}_{\mathcal{B}\sim p_\phi}
%     \left[
%         \lambda_C\frac{1}{|\mathcal{B}_A|}
%         \sum_{d\in\mathcal{B}_A}g_z(C_z(d))
%         +\lambda_D D(\mathcal{B}_A)
%     \right]
%     \quad\text{s.t.}\quad
%     \Pr_{d\sim p_\phi}[A(d)=1]\geq\alpha,
%     \label{eq:ace-objective}
% \end{equation}
\begin{tcolorbox}[
    enhanced,
    title={ACE Objective},
    colback=myblue!4,
    colbacktitle=myblue!92!black,
    colframe=myblue!88!black,
    coltitle=white,
    fonttitle=\bfseries,
    boxrule=1.1pt,
    arc=3mm,
    outer arc=3mm,
    left=7pt,
    right=7pt,
    top=7pt,
    bottom=7pt,
    toptitle=4pt,
    bottomtitle=4pt,
    lefttitle=7pt,
    righttitle=7pt,
    titlerule=0pt,
    label={box:ace-objective}
]
 A compact statement of the ACE objective is
\begin{equation}
    \max_{\phi}\;
    \mathbb{E}_{\mathcal{B}\sim p_\phi}
    \left[
        \lambda_C\frac{1}{|\mathcal{B}_A|}
        \sum_{d\in\mathcal{B}_A}g_z(C_z(d))
        +\lambda_D D(\mathcal{B}_A)
    \right]
    \quad\text{s.t.}\quad
    \Pr_{d\sim p_\phi}[A(d)=1]\geq\alpha,
    \label{eq:ace-objective}
\end{equation}
where $A(d)$ is the validity decision and $\alpha$ is a required acceptance level. If no candidate passes the accuracy gate, the utility is defined as zero. This formulation expresses the intended asymmetry: accuracy establishes the feasible set, while complexity and diversity shape which valid data are most useful.
\end{tcolorbox}

ACE therefore is an analytical lens over existing pipelines, not a partition of the literature into three disjoint method families. A verified blueprint can improve accuracy, induce a longer dependency structure, and enable controlled recombination at the same time. Conversely, a failure-driven generator may target model-relative complexity while narrowing domain coverage. The following three sections therefore decompose prior work at the mechanism level: how generation pipelines ensure accuracy, calibrate complexity, and expand diversity, including the interactions and costs among these goals.

\section{Accuracy}
\label{sec:accuracy}

Accuracy is the prerequisite of the ACE objective. In agentic data, supervision extends beyond a final answer to a coupled environment, task, interaction, and success signal. A sample can be fluent yet inaccurate because its task is infeasible, a tool is implemented inconsistently, an observation does not follow from the preceding action, or the verifier rewards the wrong outcome. Existing work addresses these failures at different points in the generation pipeline rather than through one universal accuracy method.

\subsection{Accuracy under the ACE Objective}
\label{sec:accuracy-objective}

Section~\ref{sec:ace-distribution-design} treats accuracy as an admission condition. For $d=(E,q,\tau,v)$, $A(d)$ requires the environment to be internally consistent, the task to be feasible and grounded, the interaction to respect actions and state transitions, and the verifier to match the intended outcome. Complexity and diversity are evaluated only after this gate because invalid examples cannot become useful merely by being difficult or different.

We summarize these requirements through a conjunctive validity decision and its batch-level acceptance rate:
\begin{tcolorbox}[
    enhanced,
    title={Accuracy under the ACE Objective},
    colback=myorange!4,
    colbacktitle=myorange!92!black,
    colframe=myorange!88!black,
    coltitle=white,
    fonttitle=\bfseries,
    boxrule=1.1pt,
    arc=3mm,
    outer arc=3mm,
    left=7pt,
    right=7pt,
    top=7pt,
    bottom=7pt,
    toptitle=4pt,
    bottomtitle=4pt,
    lefttitle=7pt,
    righttitle=7pt,
    titlerule=0pt,
    label={box:accuracy-objective}
]

\begin{equation}
\begin{aligned}
    A(d)
    &=V_E(E)\wedge V_q(q\mid E)\wedge
      V_\tau(\tau\mid E,q)\wedge V_v(v\mid E,q,\tau),\\
    \mathrm{Acc}(\mathcal{B})
    &=\frac{1}{|\mathcal{B}|}\sum_{d\in\mathcal{B}}\mathbb{I}[A(d)=1].
\end{aligned}
\label{eq:accuracy-definition}
\end{equation}
The conjunction is intentional: a correct-looking trajectory does not compensate for an infeasible task, and a correct terminal state does not compensate for a verifier that accepts policy-violating shortcuts.

\end{tcolorbox}
This definition combines local validity and relational consistency. Local checks concern individual components, such as a well-formed schema or executable call. Relational checks ask whether the components describe the same problem: $q$ must refer to objects and operations in $E$, observations in $\tau$ must result from executed actions, and $v$ must accept the intended terminal condition. Independently plausible components can drift apart when generated in separate stages~\citep{ivanovAnchorMitigatingArtifact2026}.

Accuracy has one conceptual standard but domain-specific evidence. Tool-use data can check schemas and API execution; coding data can compile and run tests; formal reasoning can invoke a proof assistant; web, GUI, and embodied data can inspect application or simulator state. Where no complete oracle exists, pipelines combine partial rules with semantic review. The survey question is therefore what a work verifies, when it verifies it, and which failures remain outside its verifier.

\subsection{Factor-Level Accuracy}
\label{sec:accuracy-factors}

\paragraph{Environment Accuracy.}
An accurate $E$ implements its declared tools, states, policies, and transitions consistently. Local checks test schemas, argument types, returns, and reset behavior, while global checks test coherent state changes and policy enforcement. 
When environments are directly instantiated from real-world systems, this consistency is often inherited to some extent. Therefore, the main challenge on this axis lies in simulated or synthetically constructed environments. 

Executable environments are especially valuable because they enable component-level validation before rollout and are increasingly becoming a central design choice for constructing accurate agentic environments~\citep{songEnvScalerScalingToolInteractive2026,tuScaleEnvScalingEnvironment2026,wangAgentWorldModel2026,xuEnvFactoryScalingToolUse2026}.
% It is further extended to tool-use RL and scientific discovery, where environment validity must be established before turn-level rewards are meaningful~\citep{duCodeGym2025,xuSciDisco2026}.

\paragraph{Task Accuracy.}
A valid $q$ must be interpretable and feasible under the relevant state, tools, and policies. In multi-turn settings, the task is often implicit and progressively revealed through interaction rather than fully specified upfront. The evolving intent should remain semantically coherent over time, unless it reflects an intentional change that is itself recognized and handled by the verifier. Accordingly, task validity can be enforced both before interaction, for example through task blueprints or plans~\citep{prabhakarAPIGenMTAgenticPipeline2025,wangToolFlowBoostingLLM2025}, and during interaction, through semantic consistency checks that ensure the progressively revealed intent remains coherent with the preceding context~\citep{wangTrajectory2TaskTrainingRobust2026}.

\paragraph{Interaction Accuracy.}
An accurate $\tau$ should contain schema-compliant actions, causally grounded observations, and behavior that remains consistent with the task and applicable policies.
Unlike environment and task accuracy, which primarily affect learning through the conditions they impose, the interaction trajectory directly provides the behavioral supervision from which the agent learns. Trajectory accuracy is therefore particularly critical, as errors in $\tau$ can be directly imitated by the trained model. Consequently, a substantial portion of agentic data validation focuses on trajectory-level checks, typically implemented through filtering procedures that remove structurally invalid, semantically inconsistent, or execution-inconsistent interactions~\citep{liuAPIGenAutomatedPipeline2024,liuToolACEWinningPoints2025,ouajdiExecutionFirstSynthetic2026}.

% APIGen orders format, execution, and semantic checks; ToolACE combines deterministic and model-based checking; ASTRA grounds trajectories in executable tool-graph structures~\citep{liuAPIGenAutomatedPipeline2024,liuToolACEWinningPoints2025,tianASTRAAutomatedSynthesis2026}. Execution-First anchors synthetic tool traces in actual execution, while WebSTAR and recent code-agent curation work filter trajectories at the step or trace level~\citep{ouajdiExecutionFirstSynthetic2026,heWebSTAR2025,hanTrajectoryCuration2026}. Turn-level validity is not enough when a locally valid sequence loses the original goal or violates an earlier constraint.

\paragraph{Verifier and Outcome Accuracy.}
The verifier $v$ must correctly recognize successful outcomes, reject meaningful failures, and avoid introducing unintended incentives. Database-backed systems compare states, software-engineering environments use repository setup, compilation, and tests, and formal-reasoning agents use proof-assistant acceptance~\citep{wangAgentWorldModel2026,pan2024training,jain2025r2e,yang2026swe,badertdinov2026swe,xinDeepSeekProverAdvancingTheorem2024}. 
Verifier accuracy is especially critical for RL, where $v$ directly shapes the reward signal and therefore largely determines whether optimization reinforces genuinely successful behavior or exploits flaws in the reward design.
\begin{figure*}[t]
    \centering
    \includegraphics[width=0.9\textwidth]{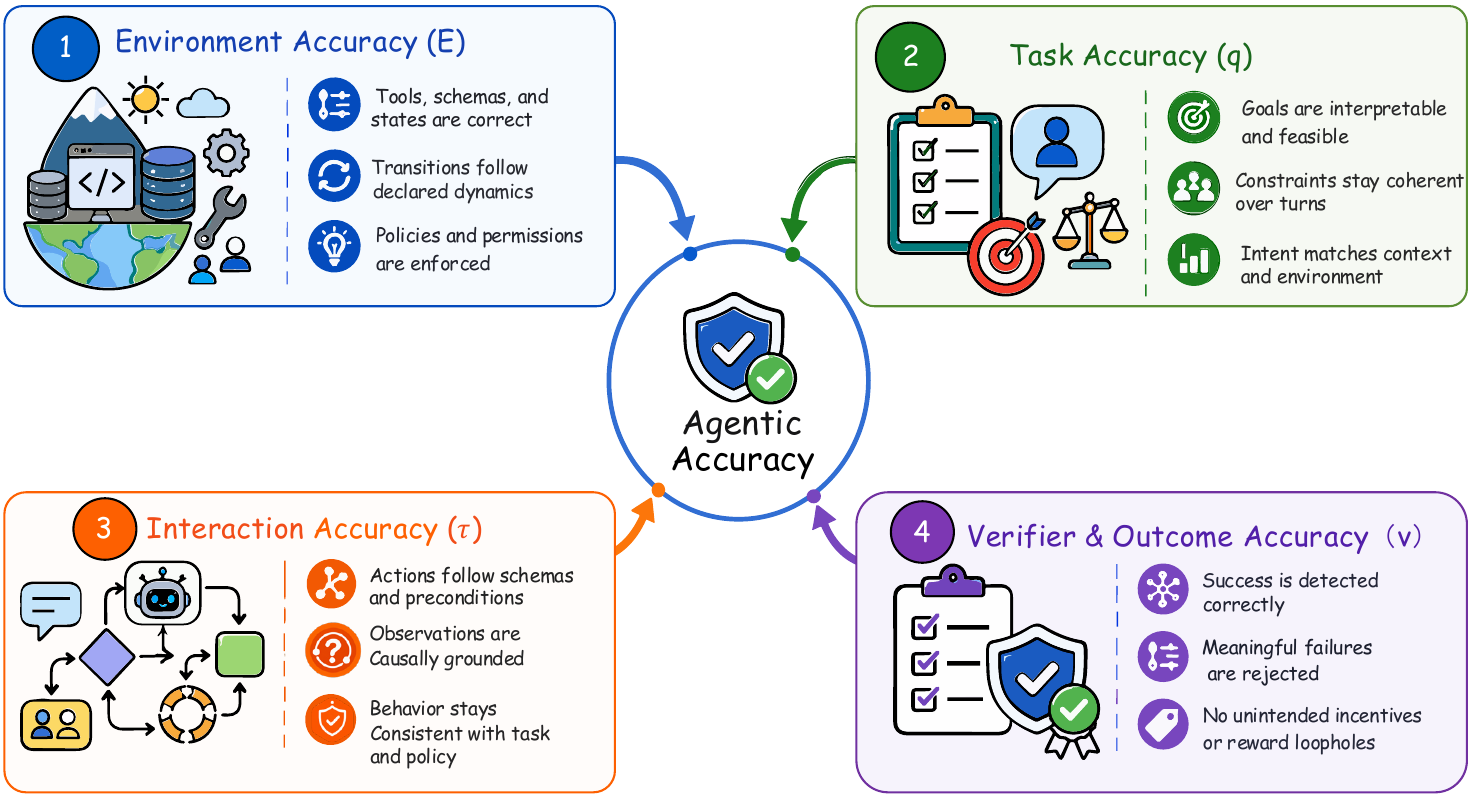}
    \caption{Four complementary factors of accuracy: environment, task, interaction, and verifier/outcome accuracy.}
    \label{fig:factor-level-accuracy}
\end{figure*}
% Search-P1 further illustrates path-centric reward design, while scientific task and environment generators couple generated problems with executable checks~\citep{xiaSearchP12026,liAutoSDT2025,xuSciDisco2026}. These signals are stronger than unconstrained preference judgments but can still miss regressions, side effects, user preferences, or reward loopholes.

\subsection{Accuracy Assurance throughout Data Generation}
\label{sec:accuracy-practices}

\begin{figure*}[t]
    \centering
    \includegraphics[width=\textwidth]{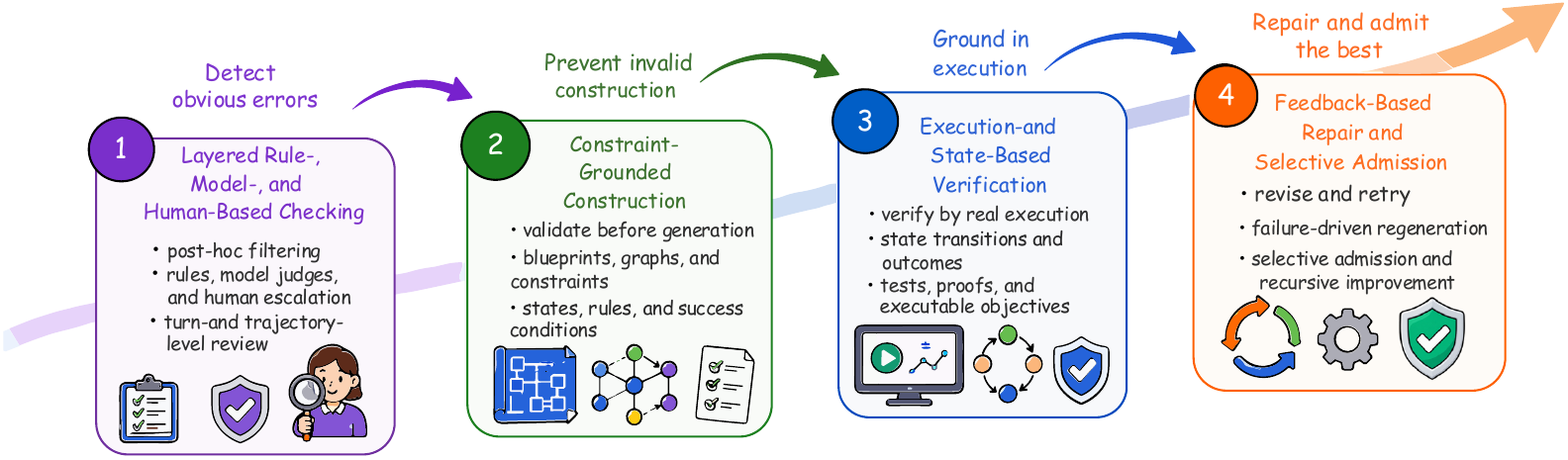}
  \caption{Four major research directions and emerging trends in accuracy assurance for agentic data generation.}
    \label{fig:accuracy-assurance}
\end{figure*}

The four components above are not mutually exclusive: a single pipeline may incorporate validity constraints during construction, execute intermediate components, and employ multiple critics or verifiers. In this subsection, we examine how existing work ensures accuracy throughout the data generation pipeline and summarize the recurring mechanisms into several high-level directions, which can be observed in Figure~\ref{fig:accuracy-assurance}.
% The literature exhibits four recurring mechanism clusters. They are not mutually exclusive method families: one pipeline may build validity into construction, execute intermediate artifacts, apply several critics, and regenerate failed components.

\subsubsection{Layered Rule-, Model-, and Human-Based Checking}

Verification and filtering are central safeguards for data accuracy, yet no single verifier can detect all structural, behavioral, and semantic failures. Existing pipelines therefore compose checks along two dimensions: \emph{cost}, from inexpensive deterministic tests to model and human review, and \emph{granularity}, from individual actions and turns to complete trajectories. APIGen~\citep{liuAPIGenAutomatedPipeline2024}, for example, applies format validation before function execution and semantic review, while ToolACE~\citep{liuToolACEWinningPoints2025} and TOUCAN~\citep{xuTOUCANSynthesizing15M2025} combine rule-based checks for schemas and arguments with model-based judgments of semantic coherence. Such ordering removes obvious errors early and reserves more expensive checks for candidates that pass objective constraints.

For long-horizon interactions, checking only the final trajectory can conceal local errors that propagate through later observations. ToolMind~\citep{yangToolMindTechnicalReport2025} and WebSTAR~\citep{heWebSTAR2025} therefore perform fine-grained turn- or step-level filtering, complementing trajectory-level assessment with localized quality signals. Other pipelines strengthen model-based verification through deliberation and cross-checking: ToolMVR~\citep{maToolMVR2025} introduces meta-verification and reflection over initial judgments, whereas APIGen-MT~\citep{prabhakarAPIGenMTAgenticPipeline2025} uses a committee of reviewers and iterative feedback to validate task blueprints before realizing full dialogues. These mechanisms reduce dependence on a single judgment, although reviewers derived from similar models may still share systematic biases.

The resulting evidence favors a layered verification stack with explicit and partially independent responsibilities. Rules establish necessary structural properties; execution and state signals test operational consistency; step- and trajectory-level critics examine causal and semantic coherence. Human review is prohibitively expensive~\citep{ramrakhyaScalingSyntheticTask2025,xuAGTTRAGENT2025}, therefore better used as a selective escalation mechanism for ambiguous goals, disagreement among automated judges, safety-sensitive outcomes, and calibration audits, particularly practical for benchmark construction~\citep{chenACEBenchWhoWins2025,heAutomaticCognitiveTask2026}.
Importantly, passing one layer should not be treated as evidence for properties that it cannot observe. Reliable filtering therefore depends not only on adding more judges, but on matching each check to a specified failure class, preserving disagreement and provenance, and escalating uncertain or high-risk samples to stronger verification.

\subsubsection{Constraint-Grounded Construction}

The second direction restricts what can be generated. APIGen-MT~\citep{prabhakarAPIGenMTAgenticPipeline2025} verifies a structured blueprint before multi-turn simulation, preventing an invalid plan from contaminating later turns. Tool graphs and workflow structures constrain candidate actions and dependencies before surface realization~\citep{wangToolFlowBoostingLLM2025,yinMagnetMultiturnTooluse2025,tianASTRAAutomatedSynthesis2026}. Stateful environment generators construct tasks together with initial states, rules, and success conditions~\citep{zengLOGIGENLogicDrivenGeneration2026,songEnvScalerScalingToolInteractive2026,tuScaleEnvScalingEnvironment2026}. Trajectory-first methods instead write tasks from paths already observed to execute~\citep{wangTrajectory2TaskTrainingRobust2026,suLearninteractDataCentricFramework2025,ouajdiExecutionFirstSynthetic2026,sunOSGenesis2024}. The shared mechanism is a structural anchor across $E$, $q$, $\tau$, and $v$.

In broader domains, coding pipelines generally start from repository state and a testable change~\citep{pan2024training,jain2025r2e,yang2026swe}; web and GUI pipelines parameterize application states and success predicates~\citep{trivediAppWorldControllable2024,rawlesAndroidWorldDynamic2025,xieOSWorldBenchmarking2024}; formal pipelines generate within a grammar or proof environment~\citep{xinDeepSeekProverAdvancingTheorem2024,alphageometry2GoldMedalist2025}.

\subsubsection{Execution- and State-Based Verification}

A growing trend is to ground verification in actual execution rather than judge whether generated data merely appear plausible. Generated actions and trajectories are executed against tools or environments, and their effects are checked through tool responses, state transitions, tests, or formal objectives. This prevents generators from fabricating observations and produces reproducible failure signals.

In tool-use and stateful environments, verification relies on real function execution, database states, and transition or terminal-state checks~\citep{liuAPIGenAutomatedPipeline2024,songEnvScalerScalingToolInteractive2026,tuScaleEnvScalingEnvironment2026,xuEnvFactoryScalingToolUse2026,wangAgentWorldModel2026,dongAgentWorldScalingRealWorld2026}. Software-engineering pipelines validate generated tasks and trajectories through repository setup, compilation, and tests~\citep{jain2025r2e,yang2026swe,badertdinov2026swe,duSWEDevFeature2025,wangSWEDevAgents2025}. Formal and scientific domains similarly use proof-assistant acceptance or executable discovery environments~\citep{xinDeepSeekProverAdvancingTheorem2024,xuSciDisco2026,liAutoSDT2025}.

Execution provides a stronger accuracy anchor than plausibility judgments alone, but it remains incomplete. A trajectory may satisfy a terminal condition while exploiting verifier loopholes, causing unintended state changes, or violating implicit constraints. Execution-based signals should therefore be combined with process checks and semantic review.

\subsubsection{Feedback-Based Repair and Selective Admission}

Verification increasingly controls subsequent generation rather than only filtering final samples. Environment builders localize failed tests and regenerate broken tools~\citep{xuEnvFactoryScalingToolUse2026}; blueprint systems revise plans before realization~\citep{prabhakarAPIGenMTAgenticPipeline2025}; rollout systems retry from the last verified state. For example, EnvFactory~\citep{xuEnvFactoryScalingToolUse2026} and EnvScaler~\citep{songEnvScalerScalingToolInteractive2026} exemplify iterative construction and testing. UI-TARS~\citep{qinUITARS2025} uses iterative trace collection and reflection, while self-evolving coding and search agents feed observed errors or outcomes back into later experience generation~\citep{xiaoSocraticSWE2026,fuSESA2026}. Rejecting or repairing a candidate at the earliest failed stage is especially valuable for long trajectories, where one invalid transition can corrupt every later observation.

Repeated repair against a fixed verifier may narrow the accepted data toward verifier-friendly patterns. Pipelines should therefore preserve alternative valid paths and use recorded failure types to guide targeted regeneration rather than merely discard failed samples~\citep{qinToolLLMFacilitatingLarge2023,haoFailureMasteryGenerating2026}.
Recursive synthesis makes this feedback loop explicit: it extends a verified seed, realigns the instruction and verifier to the longer workflow, and admits successful tasks as seeds for later rounds~\citep{liRecursiveTerminal2026}.

\subsection{Accuracy Costs, Tradeoffs, and Limitations}
\label{sec:accuracy-tradeoffs}

\paragraph{Verification Strength versus Cost.}
Rule-based checks are inexpensive, whereas execution, repository setup, simulator rollout, multi-model review, and human inspection are progressively more costly. In general, pipelines apply cheap checks early~\citep{liuAPIGenAutomatedPipeline2024,liuToolACEWinningPoints2025}, while coding pipelines automate environment setup and testing whenever possible~\citep{jain2025r2e,yang2026swe,badertdinov2026swe}. EnvFactory suggests that a smaller set of robustly verified environments can compete with simply scaling the number of environments~\citep{xuEnvFactoryScalingToolUse2026}. These findings motivate a tiered verification strategy that applies lightweight checks broadly and allocates more costly verification to samples with greater uncertainty or potential impact.

\paragraph{Verifier Coverage versus Verifier Bias.}
A verifier can only assess the properties represented in its rules or judging criteria. Deterministic checks are reliable for observable conditions but may overlook semantic errors or unintended side effects, whereas model-based judges cover less structured properties but can be inconsistent and may inherit assumptions from the generator. Meta-verification and cross-checking can reduce, but not eliminate, these errors~\citep{maToolMVR2025}. Moreover, when tasks, environments, trajectories, and verifiers are generated separately, they may remain individually plausible while becoming mutually inconsistent~\citep{ivanovAnchorMitigatingArtifact2026}. Repeated optimization against a fixed verifier can further encourage solutions that satisfy the check without fulfilling the intended task. Studies should therefore distinguish \emph{verifier acceptance} from independently audited correctness.

\paragraph{Coupling with Complexity and Diversity.}
As tasks span longer horizons, richer dependencies, and broader domains, verification must cover more possible sources of inconsistency. A sample should not be treated as difficult or novel until its feasibility and correctness have been established; otherwise, generation errors may be mistaken for useful complexity. Conversely, overly narrow verifiers may reject valid alternative strategies simply because they differ from an expected trajectory or outcome representation~\citep{prabhakarAPIGenMTAgenticPipeline2025,ivanovAnchorMitigatingArtifact2026}. Accuracy assurance must therefore expand with the complexity and diversity, while preserving consistent validity requirements across different task types and solution paths.

\paragraph{Residual Semantic Uncertainty.}

Some aspects of accuracy cannot be fully captured by executable checks or a single success condition. This uncertainty is particularly important when observations come from an LLM simulator rather than an executable environment: responses may remain locally coherent while encoding incorrect state dynamics~\citep{liSimia2025,leeEnvironmentFree2026}. A trajectory may complete the stated task while remaining inefficient, unsafe, inconsistent with user intent, or harmful in its longer-term effects. Hybrid checks can reduce this uncertainty, but cannot eliminate it. Accuracy should therefore be reported across data factors, domains, and failure categories, rather than summarized by a single aggregate pass rate.

\section{Complexity}
\label{sec:complexity}

Complexity determines whether accurate data provides an informative learning or evaluation signal. Agentic complexity is not synonymous with length, tool count, or linguistic obscurity. It is the difficulty induced by a grounded task, environment, and interaction protocol for a particular model configuration. Existing generation work manipulates structural features to create harder candidates, but increasingly relies on model behavior to decide whether those candidates are actually useful.

\subsection{Complexity under the ACE Objective}
\label{sec:complexity-objective}
\begin{tcolorbox}[
    enhanced,
    title={Complexity under the ACE Objective},
    colback=complexityblue!4,
    colbacktitle=complexityblue!92!black,
    colframe=complexityblue!88!black,
    coltitle=white,
    fonttitle=\bfseries,
    boxrule=1.1pt,
    arc=3mm,
    outer arc=3mm,
    left=7pt,
    right=7pt,
    top=7pt,
    bottom=7pt,
    toptitle=4pt,
    bottomtitle=4pt,
    lefttitle=7pt,
    righttitle=7pt,
    titlerule=0pt,
    label={box:complexity-objective}
]

For an accurate instance $d$, let $z$ denote the complete execution configuration: the target model and policy scaffold, available tools, environment protocol, verifier, sampling settings, and inference budget. We use verified failure probability as a simple model-relative definition,
\begin{equation}
    C_z(d)=1-\Pr[v(d,\tau)=1\mid d,z].
    \label{eq:model-relative-complexity}
\end{equation}
The same instance can therefore have different complexity for different models, tool access, or budgets. Structural attributes such as horizon, dependency depth, branching, partial observability, and memory demand are explanatory variables and generation controls, not universal difficulty scores. A broken environment or impossible task is excluded by the accuracy gate; its failure rate is not evidence of useful complexity.
\end{tcolorbox}

\begin{figure}[htbp]
    \centering
    \includegraphics[
        page=1,
        width=0.6\textwidth,
        keepaspectratio
    ]{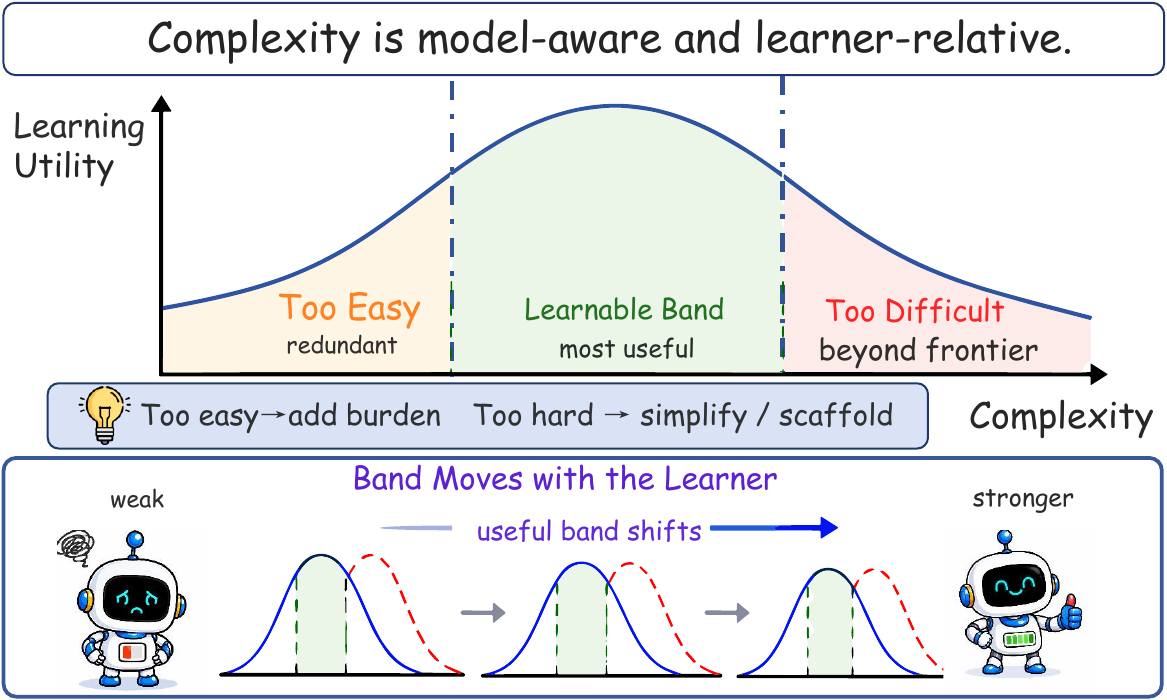}
    \caption{Learner-relative task complexity. Learning utility is maximized within a learnable band between tasks that are too easy and too difficult. This band can also shift with learner capability.}
    \label{fig:complexity}
\end{figure}
The ACE objective uses a utility over $C_z(d)$ rather than maximizing it unconditionally. Easy examples may be redundant, while examples that every available solver fails can provide no successful trajectory and may lie beyond the current learnable region. Useful data often occupies a moving band near the learner's capability frontier (see Figure~\ref{fig:complexity}). This makes complexity control bidirectional: generation can add burdens when candidates are too easy and simplify or scaffold them when they are too hard.

For comparisons between a base configuration $z_0$ and an agent-assisted configuration $z_A$, a particularly informative band contains tasks that are not reliably solved by the base model but become solvable with agentic assistance:
\begin{equation}
    p_{z_0}(d)<\rho\leq p_{z_A}(d),
    \qquad
    p_z(d)=\Pr[v(d,\tau)=1\mid d,z].
    \label{eq:agent-required-complexity}
\end{equation}
This paired criterion separates useful agent-requiring tasks from both the base-solvable tail and the beyond-frontier tail. The threshold $\rho$ is protocol-dependent rather than universal.

\subsection{Factor-Level Complexity}
\label{sec:complexity-factors}

\paragraph{Environment Complexity.}
Environment complexity is the burden induced by the actionable world in which the task is solved. It is determined by the structure of the state and action spaces, dependencies among tools or objects, transition dynamics, observation protocol, applicable policies, and the behavior of other actors~\citep{tuScaleEnvScalingEnvironment2026,luToolSandboxStateful2025,barresTau2BenchEvaluating2025}. Large environments are not necessarily complex: additional tools, records, or states matter only when they create task-relevant alternatives, prerequisites, uncertainty, or consequences. Conversely, a small action space can remain complex when actions have delayed effects, observations reveal only partial state, or repeated decisions interact through persistent state.

\paragraph{Task-signal Complexity.}
Task-signal complexity concerns what the agent must infer and what obligations a valid solution must satisfy. It can arise from compositional goals, interacting constraints, implicit or progressively revealed intent, cross-source evidence, and dependencies among subgoals~\citep{prabhakarAPIGenMTAgenticPipeline2025,shimToolDialMultiturnDialogue2025,zhangDeepPlanningBenchmarking2026}. This factor is distinct from linguistic obscurity. Missing information creates meaningful complexity only when it can be recovered from the user, environment, or available tools; otherwise, it makes the task ambiguous or infeasible. Likewise, additional conditions matter only when they jointly constrain successful behavior rather than add irrelevant detail.
\begin{figure*}[t]
    \centering
    \includegraphics[width=\textwidth]{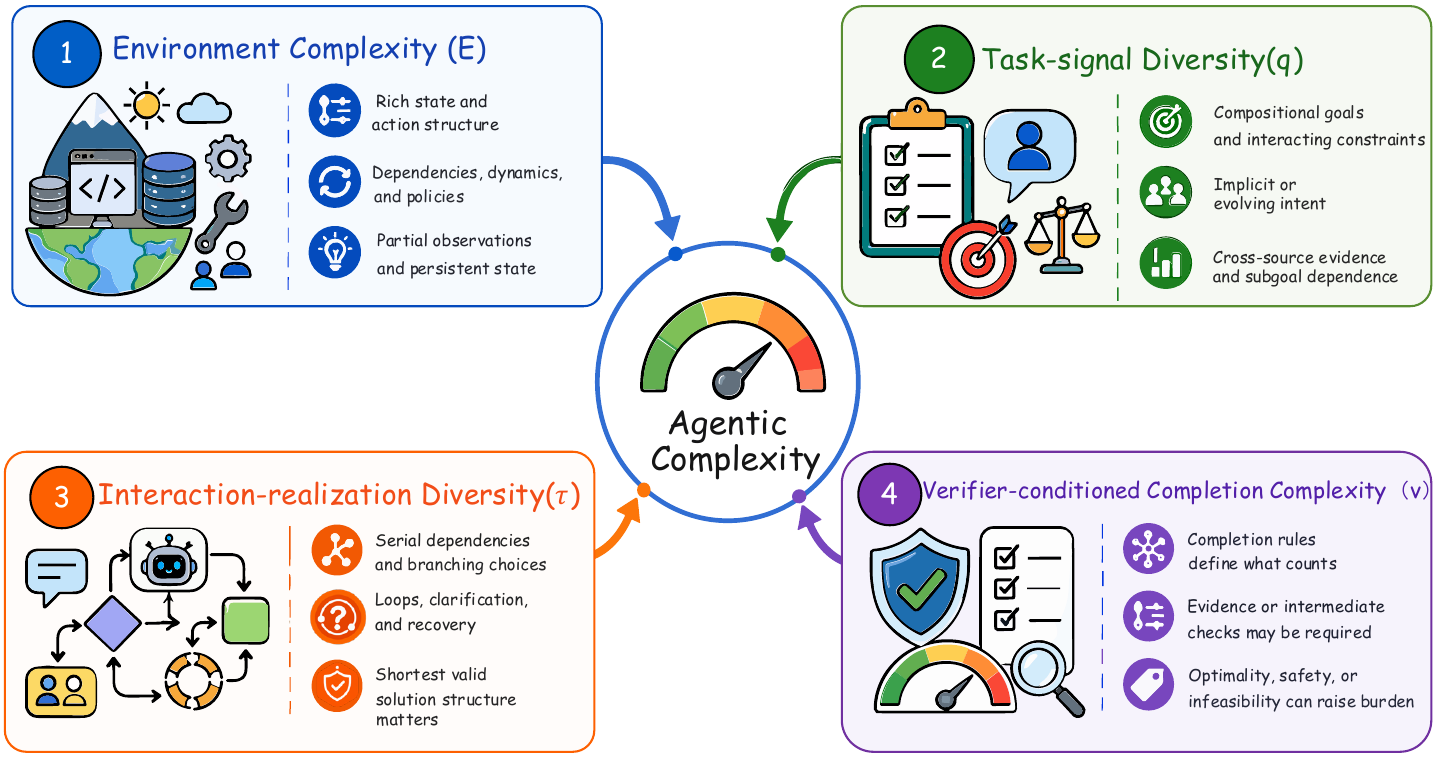}
    \caption{Factor-level complexity in agentic data across environments, tasks, interactions, and completion criteria.}
    \label{fig:factor-level-complexity}
\end{figure*}
\paragraph{Interaction-realization Complexity.}
Interaction complexity concerns about the minimum consequential interaction structure required to solve a valid task in its environment. Relevant properties include unavoidable serial dependencies, branching choices, parallel subgoals, joins, loops, clarification, and recovery from task-induced contingencies~\citep{shenTaskBenchBenchmarking2024,chenFacilitatingMultiturnFunction2025,xieAgentSynth2026}. This definition separates task-required interaction from policy-dependent inefficiency: a trajectory can be long because the problem requires many dependent steps, or because one solver repeats actions and makes mistakes. Horizon and tool-call count are therefore incomplete proxies unless they reflect the shortest valid solution structure under the stated configuration.

\paragraph{Verifier-conditioned Completion Complexity.}
The verifier influences complexity by defining what counts as completion and what evidence must be produced. A task may require any feasible terminal state, satisfaction of multiple constraints, an optimal outcome, avoidance of forbidden actions, diagnosis of infeasibility, or correctness at intermediate steps~\citep{xuCRABCrossEnvironment2025,wuMCPMarkBenchmark2026,xiaSearchP12026}. Stricter completion semantics can increase the decision burden, but only when they remain consistent with the task and observable to the verifier. Adding obligations that the verifier cannot check creates label uncertainty rather than legitimate complexity.

These factors should be interpreted jointly. The environment determines the available decision process, the task selects a goal-conditioned region of that process, the required interaction describes the dependencies that must be traversed, and the verifier specifies the accepted completion set. The same surface prompt can therefore vary in complexity through changes to any one of these components.

\subsection{Constructing and Calibrating Agentic Complexity}
\label{sec:complexity-practices}

Existing work intervenes on the factors above through several recurring mechanisms. Some methods specify dependencies before generation, while others regulate information access, alter the environment or completion conditions, transform valid seeds, or select candidates according to model behavior. These mechanisms often appear together within a broader pipeline.

\subsubsection{Structural Specification and Composition}

The most direct approach is to specify a dependency structure before realizing a task in natural language. Generators construct subgoal graphs, tool-dependency paths, scenario-skill paths, coupled constraints, or initial-to-goal plans, and then instantiate tasks that require the selected structure~\citep{yangTaskCraft2025,chenFacilitatingMultiturnFunction2025,shenTaskBenchBenchmarking2024,wangMCPBenchBenchmarking2026,fanSkillSynth2026,zhouToolVerse2026}. Related methods compose verified subtasks or action sequences so that an earlier result becomes a prerequisite for a later decision~\citep{yeToolHopQueryDriven2025,xieAgentSynth2026,yinMagnetMultiturnTooluse2025,dongAgentWorldScalingRealWorld2026}.

These structures control different burdens. Depth creates unavoidable serial dependence, width introduces parallel subgoals, joins require intermediate results to be integrated, and conditional edges make later actions depend on the current state. Similar structures appear in long-horizon navigation and branched scientific reasoning~\citep{chengOpenMobile2026,pahujaExplorer2025,zhaoAgenticIdeation2026}. However, more nodes or edges do not automatically make a task harder: an explicit chain can reveal the solution, and an added condition can prune the search space. Structural controls therefore need executable relevance and behavioral calibration.

\subsubsection{Task and Information Control}

Task-side methods adjust either what must be achieved or what information is initially available. Information can be omitted from the initial request, disclosed through later user turns, distributed across observations, or made discoverable only through tools. This creates clarification, retrieval, and state-tracking burdens in multi-turn interaction~\citep{shimToolDialMultiturnDialogue2025,zengToolACEMTNonAutoregressiveGeneration2026,wangTrajectory2TaskTrainingRobust2026,guWRITWriteRead2026}. Other methods progressively remove procedural cues or reverse-generate requests from known action sequences, producing less explicit but still grounded tasks~\citep{panditSynthesizingAgenticData2025,crouseSimulatingComplexMultiTurn2026,haoFailureMasteryGenerating2026}.

Goal-side controls instead add compositional objectives, interacting constraints, intent changes, or evidence that must be integrated across sources. Verified blueprints and reasoning graphs make these obligations explicit before dialogue realization~\citep{prabhakarAPIGenMTAgenticPipeline2025,zhangDeepPlanningBenchmarking2026,yangImpRIFStrongerImplicit2026}. Search and multimodal generators similarly vary serial or parallel decomposition and the number of evidence sources that must be reconciled~\citep{zhaoParallelSearch2025,chenGraph2EvalAutomatic2025,huangOnPolicyMultimodalSearch2026}.

The key condition is \emph{resolvability}. Withheld information must have an intended source and recovery action, while additional constraints must affect the accepted solution. Otherwise, the generator produces ambiguity or decorative detail rather than meaningful complexity.

\subsubsection{Environment and Interaction Design}

Environment-side methods change the decision process in which a task is solved. Typed tool dependencies, shared state, persistent databases, policy boundaries, and restricted observation protocols create prerequisites and consequences that span multiple actions~\citep{tuScaleEnvScalingEnvironment2026,songEnvScalerScalingToolInteractive2026,wangAgentWorldModel2026,xuEnvFactoryScalingToolUse2026}. Initial-state selection and controlled information access can further increase the amount of state that must be inferred without changing the visible goal~\citep{zengLOGIGENLogicDrivenGeneration2026,luToolSandboxStateful2025}.

The implementation is domain-dependent. Coding and scientific environments derive complexity from repository or program dependencies, executable feedback, and multi-step experimentation~\citep{duCodeGym2025,xuSciDisco2026}. Embodied environments add geometry, physics, sensing, and longer action horizons~\citep{wangEmbodiedGenV22026,zhouEmbodiedClaw2026}. Social environments introduce private information, competing incentives, communication, and strategic responses from other policies~\citep{zhouSOTOPIAInteractive2024,leiboScalableEvaluationMultiAgent2021,barresTau2BenchEvaluating2025,yashwanthSotopiaToM2026}. Across these settings, additional environment detail is useful only when it changes the transitions, observations, or choices relevant to the task.

\subsubsection{Completion and Feedback Design}

Complexity can also be controlled through the definition of success. Generated tasks may require a target state, satisfaction of coupled constraints, completion of several rubric items, avoidance of forbidden actions, or diagnosis that no feasible solution exists. State targets, constraint programs, rubric trees, and subgoal evaluators make these requirements machine-checkable~\citep{caiAutoForgeAutomatedEnvironment2025,ivanovAnchorMitigatingArtifact2026,xieQuestTrainingFrontier2026,xuCRABCrossEnvironment2025}. Programmatic benchmark generators likewise vary completion semantics across tools and environments~\citep{lyuMockWorldsReal2026,wuMCPMarkBenchmark2026}.

Feedback need not be limited to a terminal pass or fail. Path-centric rewards and step-level checks assign credit to intermediate evidence gathering, action choice, or progress toward a goal~\citep{xiaSearchP12026,heWebSTAR2025}. Such feedback can make long-horizon tasks learnable without reducing their underlying dependency structure. The task and verifier must nevertheless evolve together: adding obligations that are absent from the success check creates label error, while adding checks unrelated to the stated goal changes the benchmark rather than its complexity.

\subsubsection{Evolution and Progressive Transformation}

Instead of generating hard instances from scratch, many pipelines transform already valid seeds. Some progressively remove procedural cues or make intents less explicit~\citep{panditSynthesizingAgenticData2025,kangLearningChallenges2026}, while others strengthen goal conditions and interacting obligations~\citep{huAgentGenEnhancingPlanning2025}. Long-horizon complexity can also be increased by extending verified follow-up subtasks~\citep{xieAgentSynth2026}, composing workflow-relevant skills~\citep{fanSkillSynth2026}, or producing progressively longer workflows~\citep{liRecursiveTerminal2026}. Progressive transformation preserves a grounded core and makes the intended source of additional burden easier to identify.

Adaptive variants use execution outcomes or accumulated experience to choose subsequent transformations. They can target unresolved capabilities, derive new tasks from failures, or turn successful traces into reusable skills~\citep{zhaiAgentEvolverEfficientSelfEvolving2025,fangWebEvolver2025,fuSESA2026,xiaoSocraticSWE2026}. Because each edit can introduce infeasibility or irrelevant detail, progressive generation still requires an accuracy gate and a record of the factor being changed.

\subsubsection{Failure-Driven and Model-Aware Calibration}

The decreasing solver pass rate across verified recursive-synthesis rounds illustrates how structural growth can be checked against observed model difficulty~\citep{liRecursiveTerminal2026}.

Structural controls describe why a candidate may be hard, but solver behavior determines whether it is informative for a particular learner. Model-aware pipelines estimate verified success under a stated model, scaffold, and rollout budget, then retain candidates near a target success band or capability frontier~\citep{florensaAutomaticGoalGeneration2018,guoGenEnv2025,acikgozToolR02026,wolfBreakingSolverBottleneck2026,zeng2026toolace}. Capability profiles and observed failures can further direct generation toward specific weaknesses rather than increasing every structural feature uniformly~\citep{haoFailureMasteryGenerating2026,kangLearningChallenges2026,chenAgentFrontier2025,huangOnPolicyMultimodalSearch2026}.

Failure is informative only after validity has been established. Otherwise, frontier tasks are mixed with broken environments, impossible requests, and incomplete verifiers. Repeated verified rollouts and stratification by domain or skill are therefore needed before a candidate is prioritized, simplified, or deferred.

\subsubsection{Bidirectional Calibration and Scaffolding}

Complexity control is bidirectional. Saturated candidates can be strengthened by adding a dependency, reducing information exposure, or tightening completion requirements. Candidates beyond the useful band can be simplified by reversing those changes or by supplying intermediate goals, procedural prerequisites, action hints, and opportunities for clarification~\citep{huAgentGenEnhancingPlanning2025,liuToolACEWinningPoints2025,xuAGTTRAGENT2025,shimToolDialMultiturnDialogue2025}. Corrective feedback and reference actions can also turn otherwise inaccessible tasks into usable supervision~\citep{yinMagnetMultiturnTooluse2025,haoFailureMasteryGenerating2026}.

Additional scaffolding may divide planning, routing, specialization, and critique among auxiliary agents, while selective human assistance can resolve ambiguity or domain-specific blockers~\citep{liChainAgentsEndEndAgent2025,yaoToolACEMCPGeneralizingHistoryAware2026,jiaoAgenticProposingEnhancing2026,trinhHiLBench2026}. Because scaffolding changes the execution configuration $z$, comparisons should keep the underlying instance fixed and state whether the objective is to generate harder tasks, obtain learnable supervision, or measure the value added by assistance.

\subsection{Estimating and Reporting Complexity}
\label{sec:complexity-measurement}

Complexity evaluation should combine structural description with behavioral calibration. Structural reports should cover the manipulated factors: goal constraints, information exposure, required dependency topology, state and observation design, horizon, branching, recovery, and verifier semantics. These descriptors explain why a candidate may be hard and support controlled ablations, but they should not be collapsed into an unvalidated universal score.

Behavioral reports should state the target model, scaffold, tools, inference budget, sampling protocol, and number of rollouts. Success rates should be stratified by domain, skill, and structural band, with uncertainty where stochasticity is material. Paired evaluation is especially informative: compare matched instances before and after one intervention, or compare a base model and agent-assisted configuration on the same accepted data. Studies of code-agent trajectory curation and software-engineering data scaling further motivate reporting trajectory selection and data budget rather than treating all verified rollouts as equally informative~\citep{hanTrajectoryCuration2026,zengSkyworkSWE2025}. The reported complexity distribution should move as the learner improves rather than remain tied to a fixed teacher judgment~\citep{geAgentPsychometrics2026,guoGenEnv2025,chenAgentFrontier2025}.

The two views answer different questions. Structural descriptors support diagnosis and controllable generation, whereas behavioral estimates determine whether the resulting burden is meaningful for a learner. A convincing complexity claim therefore reports their relationship: whether deeper dependencies, reduced information, richer state, or stricter completion semantics actually lower verified success, and whether that effect persists across models rather than reflecting one solver artifact.

\subsection{Complexity Tradeoffs and Limitations}
\label{sec:complexity-tradeoffs}

\paragraph{Complexity versus Accuracy.}
Additional dependencies, state transitions, and completion requirements enlarge both the space of possible failures and the burden of verification. Long trajectories are especially fragile because an invalid early step can corrupt subsequent states. Difficulty should therefore be assessed only after schema, execution, semantic, and verifier checks; otherwise, broken environments and infeasible tasks are easily mistaken for hard cases. Strong validity constraints may reduce candidate yield, but relaxing them does not create useful complexity~\citep{jiaoAgenticProposingEnhancing2026,yangToolMindTechnicalReport2025,badertdinovSWErebenchV2LanguageAgnostic2026}.

\paragraph{Complexity versus Diversity.}
Model-aware generation often concentrates on a narrow set of current failures. This can improve short-term learning efficiency while reducing coverage of other domains, skills, solution paths, or easier examples needed for retention. Conversely, a broad mixture may contain little useful signal if most samples are already saturated. Complexity-aware selection should therefore operate within explicit coverage constraints rather than replace diversity with frontier difficulty. Fixed-budget studies showing gains from varied trajectory structures support balancing these objectives instead of optimizing either in isolation~\citep{chenBeyondQuantityTrajectory2026,chenDIVEScalingDiversity2026}.

\paragraph{Complexity versus Reality.}
Reality and complexity are related but distinct. Reality concerns whether an environment, task, and interaction plausibly reflect situations that arise in deployment, whereas complexity concerns the reasoning, information, and action burden required for completion. Realistic scenarios are often complex because they involve implicit constraints, persistent state, and noisy observations, but many real workflows remain routine; conversely, a synthetic task may be difficult yet implausible or irrelevant. Recreated environments should therefore preserve task-relevant action semantics, state transitions, and constraints rather than maximize environmental detail or difficulty for its own sake~\citep{liAgenticEnvironmentEngineering2026,chaeSafeScalableWeb2026}. Complexity claims should accordingly specify the relevant task family and deployment setting.

\paragraph{Proxy Fidelity and Calibration Cost.}
Structural proxies such as horizon, tool count, graph depth, or constraint count are inexpensive and controllable, but they do not reliably predict behavioral difficulty. Added steps may be parallelizable, extra conditions may narrow the search, and an explicit decomposition may make a longer task easier. Behavioral calibration is more informative but requires repeated executable rollouts and remains sensitive to model, scaffold, inference budget, and stochasticity~\citep{geAgentPsychometrics2026,guoGenEnv2025,wolfBreakingSolverBottleneck2026}. Practical pipelines must balance cheap structural screening against costly model-based estimates, especially near selection boundaries.

\paragraph{Difficulty versus Learnability.}
Maximizing failure rate can push generation beyond the region from which a learner receives useful supervision. All-failure candidates may provide no successful trajectory, stable reward, or evidence that the task is solvable under the available tools and budget. Frontier-oriented methods instead seek mixed outcomes or a measurable advantage from stronger scaffolds, and simplify or defer candidates that remain inaccessible~\citep{acikgozToolR02026,wolfBreakingSolverBottleneck2026,chenAgentFrontier2025}. The appropriate target also depends on use: training data should usually provide an attainable learning signal, whereas evaluation may deliberately include a harder tail to measure future progress.

\paragraph{Curriculum Drift and Closed-loop Bias.}
The useful complexity band moves as the learner improves~\citep{zeng2026toolace}. Static generators eventually saturate, while continuously adaptive generators can overfit to the current model's idiosyncratic failures, a fixed verifier, or one simulator's dynamics. Co-evolution and difficulty-aware generation address saturation, but they also make the data distribution non-stationary~\citep{guoGenEnv2025,kangLearningChallenges2026,chenAgentFrontier2025}. Held-out domains, fixed anchor sets, and periodic recalibration are needed to distinguish genuine capability growth from a generator and learner adapting to the same narrow feedback loop.

\section{Diversity}
\label{sec:diversity}

Diversity concerns the breadth of accurate and appropriately complex agentic data. It is not determined by the number of samples, prompts, tools, or domain labels, but by whether the data cover distinctions that change what an agent can observe, decide, and do. Existing pipelines pursue such coverage by expanding data sources and executable support, recombining compatible components, exploring reachable experience, constructing controlled variants, and adapting generation to remaining coverage gaps. The relevant unit is therefore a valid and behaviorally distinct environment--task--interaction relation.

\subsection{Diversity under the ACE Objective}
\label{sec:diversity-objective}

Section~\ref{sec:ace-distribution-design} treats diversity as a batch-level utility over the accuracy-filtered set $\mathcal{B}_A$. Effective diversity is \emph{validity-conditioned}: inconsistent environments and unsupported trajectories enlarge the error space rather than the useful support. It is \emph{structural}: renamed tools or paraphrased requests may still induce the same behavior. It is also \emph{learner-aware}: a distinction that is novel to one model or corpus may already be redundant for another, while unrestricted novelty may move outside the learnable region.
\begin{tcolorbox}[
    enhanced,
    title={Diversity under the ACE Objective},
    colback=diversitygreen!4,
    colbacktitle=diversitygreen!92!black,
    colframe=diversitygreen!88!black,
    coltitle=white,
    fonttitle=\bfseries,
    boxrule=1.1pt,
    arc=3mm,
    outer arc=3mm,
    left=7pt,
    right=7pt,
    top=7pt,
    bottom=7pt,
    toptitle=4pt,
    bottomtitle=4pt,
    lefttitle=7pt,
    righttitle=7pt,
    titlerule=0pt,
    label={box:diversity-objective}
]

Let $Z_E$, $Z_Q$, and $Z_I$ represent the environment-, task-, and
interaction-level factors introduced above. We define
$D(\mathcal{B})$ as the effective diversity of a generated
batch $\mathcal{B}$: the breadth of its valid and appropriately complex
factor coverage after discounting behaviorally redundant samples. A useful
abstraction is
\begin{equation}
    D(\mathcal{B})
    =
    \sum_{k\in\{E,Q,I\}} w_k
    H\!\left(
        Z_k
        \mid
        d\in\mathcal{B}_A,\,
        C_z(d)\in\mathcal{I}_z
    \right)
    -
    \lambda\,\mathrm{Red}(\mathcal{B}_A),
    \label{eq:effective-diversity}
\end{equation}
where $\mathcal{B}_A$ is the subset that passes the accuracy gate, and
$\mathcal{I}_z$ denotes the useful complexity range for learner
configuration $z$. For each factor level $k$, $H(\cdot)$ measures the
entropy of its empirical factor distribution within this filtered subset:
a larger value indicates broader and more balanced coverage.
The weight $w_k$ specifies the relative importance of environment, task,
and interaction diversity, while $\mathrm{Red}(\mathcal{B}_A)$ measures
behavioral redundancy and $\lambda$ controls its penalty. Thus, the
objective rewards balanced factor-level coverage among valid, learnable
samples rather than surface variation or raw dataset size.
\end{tcolorbox}

This formulation is a design principle rather than a universal score.
Its factor representation, entropy estimator, redundancy measure, weights,
learner configuration, and target complexity range should be specified for
each domain.

% Diversity and complexity are complementary but distinct. Complexity characterizes the difficulty of an instance under a configuration, whereas diversity characterizes variation across instances. A dataset may repeat one difficult workflow through many paraphrases, or cover many simple but behaviorally different tasks. Diversity is also not itself evidence of generalization. Coverage is a property of the data distribution; generalization is observed under a declared shift. Agents can fail differently under query, action, observation, and domain shifts, so transfer claims should identify which factor is held out~\citep{lvAgentsGeneralizeOpen2026}. Empirically, expanding tool and trajectory support can outperform adding samples from already saturated modes~\citep{chenDIVEScalingDiversity2026,chenBeyondQuantityTrajectory2026}, but these gains depend on the covered distinctions rather than volume alone.

\subsection{Factor-Level Diversity}
\label{sec:diversity-factors}

\paragraph{Environment-specification Diversity.}
This factor describes variation in the actionable world: tool and action schemas, initial states, transition dynamics, observation interfaces, policies, modalities, verifiers, and other actors. Its useful subdimensions include state, action-space, observation, dynamics, rule, and reward diversity. Environment names or tool counts are weak proxies when the resulting systems expose the same preconditions, effects, information paths, and success conditions. Stateful tool and application environments illustrate why changing a permission, policy, or hidden state may create more consequential variation than adding another nominal API~\citep{luToolSandboxStateful2025,trivediAppWorldControllable2024,barresTau2BenchEvaluating2025}.

\paragraph{Task-signal Diversity.}
Task diversity concerns what the agent is asked or expected to achieve. It spans goals and capabilities, interacting constraints, user profiles, intent revelation, feasibility regimes, temporal conditions, and difficulty bands. Surface requests are meaningfully different only when they alter the evidence, decision, or acceptable outcome. In particular, clarification, refusal, partial completion, recovery, and successful completion should be treated as distinct task regimes rather than collapsed into generic hard cases~\citep{shimToolDialMultiturnDialogue2025,wangTrajectory2TaskTrainingRobust2026}. Difficulty is one task factor, but a broad difficulty distribution can still be narrow in goals, users, or required capabilities.
\begin{figure*}[t]
    \centering
    \includegraphics[width=\textwidth]{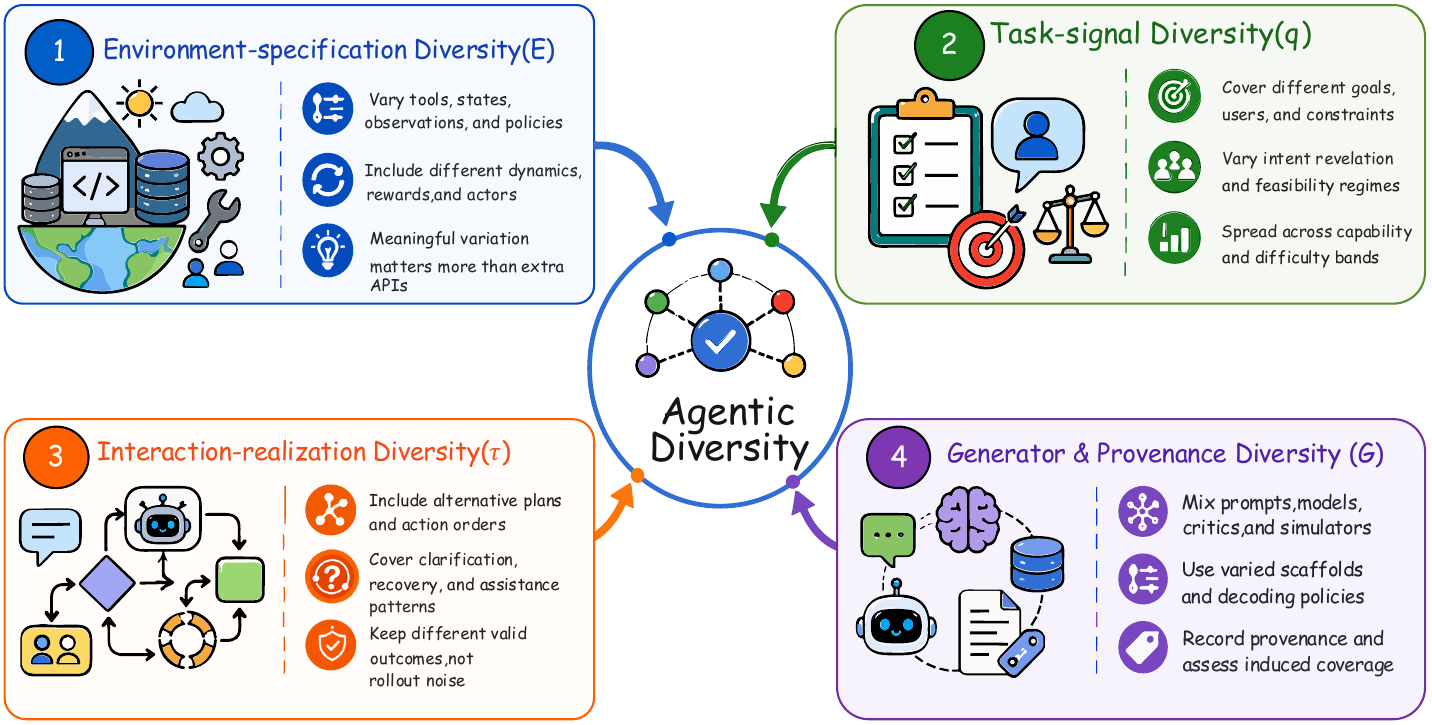}
   \caption{Four complementary factors of diversity: environment, task, interaction, and generator/provenance.}
    \label{fig:factor-level-diversity}
\end{figure*}
\paragraph{Interaction-realization Diversity.}
For a fixed environment and task, several valid interaction processes may exist. This factor covers horizon and turn structure, alternative plans and action orders, tool substitutions, action representations, clarification and recovery patterns, assistance, and outcomes. The distinction is behavioral: alternative paths that reach valid states reveal strategy coverage, whereas repeated hesitation or accidental detours are rollout noise. Failure traces contribute only when the failure is grounded and the correction or outcome is verified~\citep{haoFailureMasteryGenerating2026,gaoSelfEvolvingSyntheticData2026}.

\paragraph{Generator and provenance Diversity.}
Generated data also inherit the priors of source corpora, teacher models, prompts, simulators, critics, decoding policies, and agent scaffolds. Mixing these sources can reduce a single generator's blind spots, but source heterogeneity is an intervention rather than an outcome: different model names may still reproduce the same task and trajectory templates. Provenance should therefore be recorded and its value assessed through the induced factor coverage and learner utility~\citep{mitraAgentInstructGenerativeTeaching2024,xuTOUCANSynthesizing15M2025,yangUnderstandingAgentScaling2026}.

\subsection{Expanding Agentic Diversity: Mechanisms and Domain Evidence}
\label{sec:diversity-practices}

Prior work rarely optimizes diversity as an isolated objective. Instead, broader generation pipelines use several recurring mechanisms to enlarge or rebalance one or more of the factors above.

\subsubsection{Source and Support Expansion}

The most direct mechanism adds environments, tools, repositories, scenes, users, modalities, task sources, or verification regimes. Tool-oriented work scales real and synthetic API ecosystems, software-agent pipelines broaden executable repositories and languages, and embodied or scientific systems construct simulation-ready worlds and discovery spaces~\citep{qinToolLLMFacilitatingLarge2023,xuTOUCANSynthesizing15M2025,badertdinovSWErebenchV2LanguageAgnostic2026,duCodeGym2025,wangEmbodiedGenV22026,xuSciDisco2026}. Heterogeneous documents and real traces offer authentic long-tail support but inherit source imbalance; model generation is more controllable but can repeat teacher priors; programmatic construction supports execution and reset but may reproduce a small set of templates.

Nominal scale is therefore an incomplete measure. Broader environment distributions can improve transfer~\citep{tuScaleEnvScalingEnvironment2026,dongAgentWorldScalingRealWorld2026}, yet a smaller collection of robustly verified environments can be more useful than a larger but unreliable one~\citep{xuEnvFactoryScalingToolUse2026}. What matters is the marginal addition of a valid capability, relation, state transition, or feedback structure.

\subsubsection{Compositional Recombination}

Composition creates new relations among existing components: multi-tool dependencies, cross-application workflows, interacting user constraints, cross-file edits, or multi-object tasks. Graph- and blueprint-based pipelines sample compatible components before realizing a task or trajectory, reducing contradictions that arise from independent combination~\citep{wangToolFlowBoostingLLM2025,yinMagnetMultiturnTooluse2025,prabhakarAPIGenMTAgenticPipeline2025}. Related methods recombine task graphs, capabilities, and subgoals across tool use, computer use, coding, and robotics~\citep{chenFacilitatingMultiturnFunction2025,xieAgentSynth2026,heAutomaticCognitiveTask2026,nasirianyRoboCasaLargeScaleSimulation2024}. Composition adds diversity only when the components interact and change the required behavior; concatenating independent subtasks mainly increases length.

\subsubsection{Exploration and Experience-First Discovery}

Rather than writing a task first, experience-first methods traverse an environment, simulator, repository, proof system, or hypothesis space and derive tasks or training signals from reachable behavior. This pattern supports tool and GUI interaction, robotics, program discovery, and scientific search~\citep{ramrakhyaScalingSyntheticTask2025,suLearninteractDataCentricFramework2025,wangRoboGenTowardsUnleashing2023,romera-paredesMathematicalDiscoveriesProgram2024,zhaoAgenticIdeation2026}. Reverse generation from executed tool combinations further grounds tasks in observed evidence~\citep{chenDIVEScalingDiversity2026}, while exploration-driven GUI pipelines broaden states, applications, and recovery paths~\citep{pahujaExplorer2025,sunOSGenesis2024,chengOpenMobile2026}.

Reachability does not guarantee broad support: a narrow explorer repeatedly visits familiar states and strategies, and reward-driven search can ignore useful but hard-to-score behavior. Multiple policies, external seeds, resets, novelty signals, and state-coverage estimates help separate what an environment supports from what the current explorer happened to discover.

\subsubsection{Perturbation and Counterfactual Variation}

Controlled variation changes selected factors while holding others fixed, including initial state, policy, user constraint, observation format, tool semantics, layout, or physical parameters. Such perturbations are used to expose specific query, action, observation, and domain shifts~\citep{lvAgentsGeneralizeOpen2026}; domain randomization similarly broadens visual and physical conditions for embodied transfer~\citep{tobinDomainRandomizationTransferring2017,mehtaActiveDomainRandomization2020,chenRoboTwin2025}. Matched counterfactuals are especially informative because they reveal whether a change in state or policy causes the intended change in action. Variants that alter only surface form add apparent entropy, while implausible perturbations leave the deployment distribution; both require validity and behavioral checks.

\subsubsection{Coverage-Guided Balancing and Adaptation}

Coverage-guided pipelines target under-represented factor combinations, remove redundancy, or allocate generation toward named transfer gaps. Model-adaptive pipelines instead use failures, uncertainty, or a solve-rate band to produce experience near the learner's current frontier~\citep{haoFailureMasteryGenerating2026,chenAgentFrontier2025,guoGenEnv2025,kangLearningChallenges2026,wolfBreakingSolverBottleneck2026}. Self-evolving systems feed observed behavior, accumulated experience, or world-model errors back into task and experience generation~\citep{zhaiAgentEvolverEfficientSelfEvolving2025,fangWebEvolver2025,fuSESA2026,xiaoSocraticSWE2026}, while trajectory-aware allocation explicitly targets under-covered reasoning and interaction structures~\citep{chenBeyondQuantityTrajectory2026}.

Adaptation improves marginal efficiency but can overfit generation to transient model failures. Coverage constraints, stratified replay, and periodic broad exploration are needed to preserve skills, domains, and easier anchor cases while the frontier moves.
\begin{figure*}[t]
    \centering
    \includegraphics[width=\textwidth]{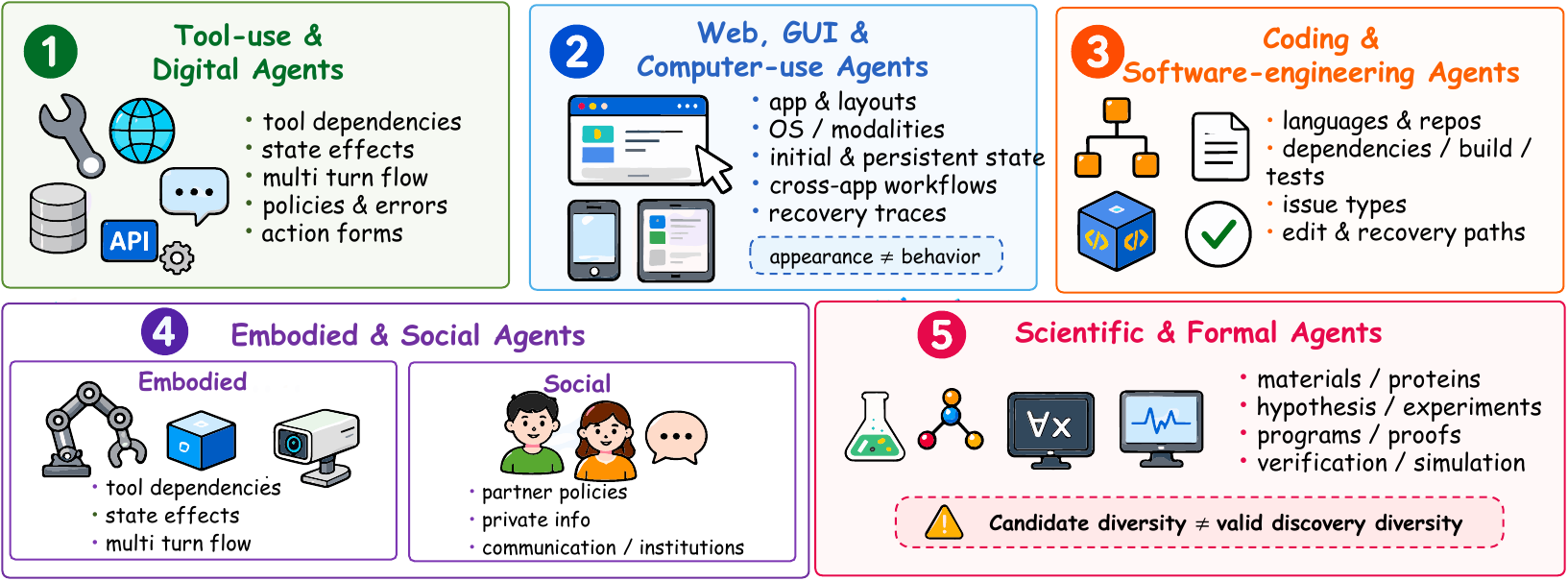}
    \caption{
        Domain-specific instantiations of diversity across
        diverse kinds of agents.
        Meaningful diversity should be characterized by
        domain-specific behavioral units rather than surface-level
        variation alone.
    }
    \label{fig:domain-specific-diversity}
\end{figure*}
% ============================================================
% Required packages
% ============================================================

% \usepackage{booktabs}
% \usepackage{tabularx}
% \usepackage{array}
% \usepackage[table]{xcolor}
% \usepackage{hyperref}
% \usepackage{fontawesome5}
% \usepackage{caption}

% ============================================================
% Colors
% ============================================================

\definecolor{groupblue}{RGB}{214,232,248}
\definecolor{githubpink}{RGB}{230,45,110}

% ============================================================
% GitHub link
% ============================================================

% \newcommand{\githubrepo}[1]{%
%   \href{#1}{%
%     \textcolor{black}{\faGithub}\,
%     \textcolor{githubpink}{GitHub}%
%   }%
% }

% ============================================================
% Float settings
% ============================================================

\renewcommand{\topfraction}{0.95}
\renewcommand{\dbltopfraction}{0.95}
\renewcommand{\textfraction}{0.05}
\renewcommand{\floatpagefraction}{0.90}
\renewcommand{\dblfloatpagefraction}{0.90}

% ============================================================
% Domain-Specific Diversity Table -- Part I
% ============================================================

\begin{table*}[!t]
\centering

\caption{
\textbf{Representative works on domain-specific diversity in agentic data generation.}
Methods are grouped by their primary application domain.
\textbf{Key Mechanism} summarizes the domain-specific source of meaningful
behavioral diversity, while \textbf{Resource} links to public GitHub
repositories or Hugging Face websites when available.
}
\label{tab:domain-specific-diversity}

% ------------------------------------------------------------
% Overall table style
% ------------------------------------------------------------

\footnotesize
\setlength{\tabcolsep}{3.2pt}
\renewcommand{\arraystretch}{1.08}

% ============================================================
% Column widths
%
% Method          24%
% Domain          12%
% Key Mechanism   remaining width
% Resource        9.5%
% ============================================================

\begin{tabularx}{\linewidth}{@{}
>{\raggedright\arraybackslash}p{0.24\linewidth}
>{\raggedright\arraybackslash}p{0.12\linewidth}
>{\raggedright\arraybackslash}X
>{\centering\arraybackslash}p{0.095\linewidth}
@{}}

\toprule

\textbf{Method}
&
\textbf{Domain}
&
\textbf{Key Mechanism}
&
\textbf{Resource}
\\

\midrule

% ============================================================
% Tool-use and Digital Agents
% ============================================================

\rowcolor{groupblue}
\multicolumn{4}{@{}l@{}}{%
  \rule{0pt}{2.15ex}\textbf{Tool-use and Digital Agents}
}
\\

ToolLLM~\citep{qinToolLLMFacilitatingLarge2023}
&
Tool Use
&
Large-scale coverage of real-world APIs and tool capabilities
&
\githubrepo{https://github.com/OpenBMB/ToolBench}
\\

APIGen~\citep{liuAPIGenAutomatedPipeline2024}
&
Tool Use
&
Verifiable function-calling generation across diverse APIs, tasks, and executions
&
\githubrepo{https://github.com/SalesforceAIResearch/xLAM}
\\

ToolACE~\citep{liuToolACEWinningPoints2025}
&
Tool Use
&
Diversification of tool pools, function-calling tasks, and action representations
&
\huggingfacerepo{https://huggingface.co/datasets/Team-ACE/ToolACE}
\\

ToolFlow~\citep{wangToolFlowBoostingLLM2025}
&
Tool Use
&
Coherent tool dependencies and multi-turn information flow through structured synthesis
&
--
\\

ACE-Router~\citep{yaoToolACEMCPGeneralizingHistoryAware2026}
&
Tool / MCP
&
History-aware routing across heterogeneous MCP tools and interaction histories
&
\githubrepo{https://github.com/euyis1019/ACE-Router}
\\

DeepResearcher~\citep{zhengDeepResearcher2025}
&
Search
&
Diverse real-world research trajectories, retrieval behavior, and evidence integration
&
\githubrepo{https://github.com/GAIR-NLP/DeepResearcher}
\\

WebSailor-V2~\citep{liWebSailorV22025}
&
Search
&
Synthetic deep-search experience with diverse retrieval and evidence paths
&
\githubrepo{https://github.com/Alibaba-NLP/DeepResearch}
\\

Search-R1~\citep{jinSearchR12025}
&
Search
&
Reinforcement learning over interleaved reasoning and search-engine interaction
&
\githubrepo{https://github.com/PeterGriffinJin/Search-R1}
\\

ParallelSearch~\citep{zhaoParallelSearch2025}
&
Search
&
Query decomposition and parallelized sub-query search trajectories
&
--
\\

DIVE~\citep{chenDIVEScalingDiversity2026}
&
Tool Use
&
Expansion of tool-pool and per-task toolset coverage for OOD generalization
&
\githubrepo{https://github.com/sheep333c/DIVE}
\\

% ============================================================
% Web, GUI, and Computer-use Agents
% ============================================================

\addlinespace[4pt]
\midrule

\rowcolor{groupblue}
\multicolumn{4}{@{}l@{}}{%
  \rule{0pt}{2.15ex}\textbf{Web, GUI, and Computer-use Agents}
}
\\

OSWorld~\citep{xieOSWorldBenchmarking2024}
&
Computer Use
&
Coverage of real computer environments, applications, states, and open-ended workflows
&
\githubrepo{https://github.com/xlang-ai/OSWorld}
\\

AndroidWorld~\citep{rawlesAndroidWorldDynamic2025}
&
Mobile / GUI
&
Dynamic Android applications with state-dependent task instantiation
&
\githubrepo{https://github.com/google-research/android_world}
\\

CRAB~\citep{xuCRABCrossEnvironment2025}
&
GUI
&
Cross-environment multimodal tasks spanning heterogeneous applications and interfaces
&
\githubrepo{https://github.com/camel-ai/crab}
\\

Scaling Synthetic Task Generation
~\citep{ramrakhyaScalingSyntheticTask2025}
&
Computer Use
&
Exploration-driven synthesis that expands reachable states, tasks, and interaction paths
&
--
\\

OS-Genesis~\citep{sunOSGenesis2024}
&
GUI
&
Reverse task synthesis from GUI trajectories and executable interaction paths
&
\githubrepo{https://github.com/OS-Copilot/OS-Genesis}
\\

OpenMobile~\citep{chengOpenMobile2026}
&
Mobile
&
Task and trajectory generation across applications and interaction states
&
\githubrepo{https://github.com/njucckevin/OpenMobile-Code}
\\

% ============================================================
% Coding and Software-engineering Agents
% ============================================================

\addlinespace[4pt]
\midrule

\rowcolor{groupblue}
\multicolumn{4}{@{}l@{}}{%
  \rule{0pt}{2.15ex}\textbf{Coding and Software-engineering Agents}
}
\\

SWE-rebench V2
~\citep{badertdinovSWErebenchV2LanguageAgnostic2026}
&
Software Eng.
&
Language-agnostic scaling of executable software-engineering tasks and repositories
&
\githubrepo{https://github.com/SWE-rebench/SWE-rebench-V2}
\\

Open-SWE-Traces~\citep{ahmadOpenSWETraces2026}
&
Software Eng.
&
Multilingual software-engineering trajectories with diverse solution and feedback paths
&
--
\\

SWE-Dev~\citep{duSWEDevFeature2025}
&
Software Eng.
&
Feature-level software-development tasks coupled with executable tests
&
\githubrepo{https://github.com/DorothyDUUU/SWE-Dev}
\\

SWE-dev~\citep{wangSWEDevAgents2025}
&
Software Eng.
&
Training trajectories spanning repositories, task types, edit scopes, and execution feedback
&
\githubrepo{https://github.com/THUDM/SWE-Dev}
\\

Beyond Quantity
~\citep{chenBeyondQuantityTrajectory2026}
&
Coding
&
Fixed-budget allocation across distinct trajectory structures rather than repeated samples
&
--
\\

\bottomrule

\end{tabularx}

\end{table*}

% ============================================================
% Domain-Specific Diversity Table -- Part II
% ============================================================

\begin{table*}[!t]
\ContinuedFloat
\centering

\caption{
\textbf{Representative works on domain-specific diversity in agentic data generation (continued).}
}

\footnotesize
\setlength{\tabcolsep}{3.2pt}
\renewcommand{\arraystretch}{1.08}

% ============================================================
% Same column widths as Part I
% ============================================================

\begin{tabularx}{\linewidth}{@{}
>{\raggedright\arraybackslash}p{0.24\linewidth}
>{\raggedright\arraybackslash}p{0.12\linewidth}
>{\raggedright\arraybackslash}X
>{\centering\arraybackslash}p{0.095\linewidth}
@{}}

\toprule

\textbf{Method}
&
\textbf{Domain}
&
\textbf{Key Mechanism}
&
\textbf{Resource}
\\

\midrule

% ============================================================
% Embodied and Social Agents
% ============================================================

\rowcolor{groupblue}
\multicolumn{4}{@{}l@{}}{%
  \rule{0pt}{2.15ex}\textbf{Embodied and Social Agents}
}
\\

ProcTHOR~\citep{procthor2022}
&
Embodied
&
Procedural generation of diverse scenes, objects, layouts, and embodied environments
&
\githubrepo{https://github.com/allenai/procthor}
\\

Holodeck~\citep{yangHolodeckLanguageGuided2023}
&
Embodied
&
Language-guided generation of interactive 3D environments and scene configurations
&
\githubrepo{https://github.com/allenai/Holodeck}
\\

RoboCasa~\citep{nasirianyRoboCasaLargeScaleSimulation2024}
&
Embodied
&
Large-scale simulation of everyday scenes, objects, affordances, and manipulation tasks
&
\githubrepo{https://github.com/robocasa/robocasa}
\\

Automatic Cognitive Task Generation
~\citep{heAutomaticCognitiveTask2026}
&
Embodied
&
Generation of activities and task structures under physical and environmental constraints
&
--
\\

EmbodiedGen V2~\citep{wangEmbodiedGenV22026}
&
Embodied
&
Simulation-ready 3D world generation with diverse scenes and physical interaction conditions
&
\githubrepo{https://github.com/HorizonRobotics/EmbodiedGen}
\\

SOTOPIA~\citep{zhouSOTOPIAInteractive2024}
&
Social
&
Open-ended social interactions with diverse roles, goals, and partner behavior
&
\githubrepo{https://github.com/sotopia-lab/sotopia}
\\

Melting Pot~\citep{leiboScalableEvaluationMultiAgent2021}
&
Multi-Agent
&
Diverse partner policies, incentives, social situations, and interaction structures
&
\githubrepo{https://github.com/google-deepmind/meltingpot}
\\

Concordia~\citep{vezhnevetsGenerativeAgentBased2023}
&
Social
&
Generative agent-based simulation grounded in physical, social, or digital contexts
&
\githubrepo{https://github.com/google-deepmind/concordia}
\\

SOTOPIA-ToM~\citep{yashwanthSotopiaToM2026}
&
Social
&
Diverse private-information settings and information-management demands in multi-agent interaction
&
--
\\

SynthAgent~\citep{aghaeeSynthAgent2026}
&
Social / Simulation
&
Persona-conditioned longitudinal behavior and outcomes grounded in heterogeneous evidence
&
--
\\

% ============================================================
% Scientific and Formal Agents
% ============================================================

\addlinespace[4pt]
\midrule

\rowcolor{groupblue}
\multicolumn{4}{@{}l@{}}{%
  \rule{0pt}{2.15ex}\textbf{Scientific and Formal Agents}
}
\\

Scaling Deep Learning for Materials Discovery
~\citep{merchantScalingDeepLearning2023}
&
Materials
&
Large-scale exploration of candidate materials under learned property prediction
&
\githubrepo{https://github.com/google-deepmind/materials_discovery}
\\

RFdiffusion~\citep{watsonDenovoDesignProtein2023}
&
Protein Design
&
Generative exploration of diverse protein structures and functions
&
\githubrepo{https://github.com/RosettaCommons/RFdiffusion}
\\

SciDisco~\citep{xuSciDisco2026}
&
Scientific
&
Diverse scientific-discovery processes in executable environments with process-level verification
&
--
\\

AutoSDT~\citep{liAutoSDT2025}
&
Scientific
&
Generation of executable data-driven scientific discovery tasks across heterogeneous datasets
&
\githubrepo{https://github.com/OSU-NLP-Group/AutoSDT}
\\

FunSearch~\citep{romera-paredesMathematicalDiscoveriesProgram2024}
&
Formal / Math
&
Exploration of executable programs under automated evaluation and iterative program search
&
\githubrepo{https://github.com/google-deepmind/funsearch}
\\

DeepSeek-Prover~\citep{xinDeepSeekProverAdvancingTheorem2024}
&
Formal
&
Large-scale synthetic theorem generation with proof-assistant-based formal verification
&
\githubrepo{https://github.com/deepseek-ai/DeepSeek-Prover-V1.5}
\\

\bottomrule

\end{tabularx}

\end{table*}
\subsubsection{Domain-Specific Instantiations and Evidence}
\label{sec:diversity-domains}

The mechanisms are shared, but meaningful factors and evidence differ by domain. Within-domain diversity covers states, workflows, users, strategies, and outcomes inside one deployment family; cross-domain diversity covers qualitatively different semantics, interfaces, dynamics, and verification regimes. 
% Domain labels alone establish neither.
Figure~\ref{fig:domain-specific-diversity} summarizes the domain-specific instantiations of diversity across diverse kinds of agents and Table~\ref{tab:domain-specific-diversity} lists the representative works. Below explores the details.

\paragraph{Tool-use and Digital Agents.}
Coverage initially centered on API and capability pools, but later pipelines diversify executable state effects, coherent tool dependencies, multi-turn information flow, policies, errors, and action representations~\citep{qinToolLLMFacilitatingLarge2023,liuAPIGenAutomatedPipeline2024,liuToolACEWinningPoints2025,wangToolFlowBoostingLLM2025,yaoToolACEMCPGeneralizingHistoryAware2026}. Search agents add variation in sources, retrieval engines, decomposition, evidence paths, and query revision even when the nominal search interface is unchanged~\citep{zhengDeepResearcher2025,liWebSailorV22025,jinSearchR12025,zhaoParallelSearch2025}. Tool identity thus substantially undercounts behavioral support. Controlled tool-use studies further show that tool-pool and per-task toolset coverage can improve out-of-distribution performance more efficiently than repeated sampling from the same support~\citep{chenDIVEScalingDiversity2026}.

\paragraph{Web, GUI, and Computer-use Agents.}
Relevant variation spans applications, layouts, operating systems, modalities, initial states, persistent effects, and cross-application workflows. Existing work combines real or recreated interfaces with procedural task generation, reverse synthesis, environment exploration, multimodal actions, and recovery traces~\citep{xieOSWorldBenchmarking2024,rawlesAndroidWorldDynamic2025,xuCRABCrossEnvironment2025,ramrakhyaScalingSyntheticTask2025,sunOSGenesis2024,chengOpenMobile2026}. This domain separates visual from behavioral diversity particularly clearly: very different screenshots may share one interaction topology, while a small permission or state change on the same screen may reverse the correct action. Transfer to unseen real interfaces is therefore stronger evidence than generated page count.

\paragraph{Coding and Software-engineering Agents.}
Environment coverage includes languages, repository structures, dependencies, build systems, tests, and project histories; task and trajectory coverage include issue types, localization paths, edit scopes, test feedback, and recovery. Recent pipelines scale executable repositories and multilingual trajectories, generate feature-level tasks and tests, and allocate fixed budgets across distinct trajectory structures~\citep{badertdinovSWErebenchV2LanguageAgnostic2026,ahmadOpenSWETraces2026,duSWEDevFeature2025,wangSWEDevAgents2025,chenBeyondQuantityTrajectory2026}. The meaningful unit is a verified problem--repository--feedback relation, not an issue description or repository count.

\paragraph{Embodied and Social Agents.}

Embodied diversity spans scenes, objects, affordances, physics, sensors, embodiments, and action effects. Procedural world generation becomes behaviorally useful when scene variation is coupled with activities, physical constraints, task graphs, or policy changes~\citep{procthor2022,yangHolodeckLanguageGuided2023,nasirianyRoboCasaLargeScaleSimulation2024,heAutomaticCognitiveTask2026,wangEmbodiedGenV22026}. Social environments add partner policies, incentives, private information, communication channels, and institutions; persona variation alone is weak unless these factors change transitions or rewards~\citep{zhouSOTOPIAInteractive2024,leiboScalableEvaluationMultiAgent2021,vezhnevetsGenerativeAgentBased2023,yashwanthSotopiaToM2026}. Domain-specific simulations can further couple personas with longitudinal behavior and outcomes, as in virtual patients grounded in clinical, behavioral, and psychosocial sources~\citep{aghaeeSynthAgent2026}.

\paragraph{Scientific and Formal Agents.}
Here, candidate breadth is tightly constrained by feasibility and verification. Generative systems explore materials, proteins, hypotheses, experiments, programs, and proofs under property predictors, execution, simulation, or formal checking~\citep{merchantScalingDeepLearning2023,watsonDenovoDesignProtein2023,xuSciDisco2026,liAutoSDT2025,romera-paredesMathematicalDiscoveriesProgram2024,xinDeepSeekProverAdvancingTheorem2024}. These domains make the distinction between candidate diversity and valid discovery diversity explicit: verifier coverage must expand with the candidate space, or the apparently broader support cannot be trusted.

\subsection{Measuring Diversity}
\label{sec:diversity-measurement}

Agentic diversity cannot be summarized reliably by a single number. Its measurement should proceed in stages: first determine which environment, task, and interaction factors are covered; then check whether the resulting samples require genuinely different behavior, remain valid and appropriately difficult, transfer to held-out settings, and provide additional value to the learner. Each study should therefore state explicitly which differences it regards as meaningful.

\paragraph{Factor Coverage and Balance.}
The most direct measurement is to divide each factor level into a set of categories. For example, task diversity may be partitioned into information seeking, transaction, clarification, refusal, recovery, and so on, while environment diversity may be partitioned by tool family, state regime, or policy type. 
\begin{tcolorbox}[
enhanced jigsaw,
breakable,
    title={Factor Coverage and Balance},
    colback=myblue!4,
    colbacktitle=myblue!92!black,
    colframe=myblue!88!black,
    coltitle=white,
    fonttitle=\bfseries,
    boxrule=1.1pt,
    arc=3mm,
    outer arc=3mm,
    left=7pt,
    right=7pt,
    top=7pt,
    bottom=7pt,
    toptitle=4pt,
    bottomtitle=4pt,
    lefttitle=7pt,
    righttitle=7pt,
    titlerule=0pt,
    label={box:diversity-coverage}
]
Let $\mathcal{S}_k$ denote the declared categories for factor level $k\in\{E,Q,I\}$, let $n_b$ be the number of valid samples in category $b$, and let $\widehat{p}_b$ be its empirical proportion. Coverage and normalized entropy then provide two simple summaries:
\begin{equation}
    \mathrm{Cov}_k
    =\frac{|\{b\in\mathcal{S}_k:n_b>0\}|}{|\mathcal{S}_k|},
    \qquad
    \widetilde{H}_k
    =-\frac{1}{\log|\mathcal{S}_k|}
      \sum_{b\in\mathcal{S}_k}\widehat{p}_b\log\widehat{p}_b.
    \label{eq:diversity-coverage}
\end{equation}
Here, $\mathrm{Cov}_k$ is the fraction of declared categories represented by at least one sample. The normalized entropy $\widetilde{H}_k$ measures how evenly samples are distributed across those categories: it reaches one when the distribution is uniform and decreases as the data concentrate in a few categories. The two statistics are complementary. Coverage can be high even when nearly all samples belong to one dominant category, whereas entropy alone does not show which important categories are absent.
\end{tcolorbox}
Marginal statistics should also be supplemented with a small number of meaningful joint distributions, such as tool family $\times$ task capability or state regime $\times$ required outcome. Otherwise, each factor may appear diverse on its own while occurring only in fixed combinations. This ``template locking'' creates nominal coverage without broadening the relations that agents must learn. Full Cartesian coverage is rarely practical, so studies should select combinations that correspond to expected deployment shifts or known dependencies.

\paragraph{Behavioral Non-redundancy.}
Category coverage still overestimates diversity when differently worded samples require essentially the same behavior. Text similarity can detect paraphrases, but agentic data should also be compared through tool-call or action graphs, state transitions, dependency paths, clarification and recovery patterns, and final-state changes. These representations help distinguish alternative strategies from renamed tools, rewritten requests, or inconsequential rollout variation. Graph fingerprints and kernel-based effective-number measures can summarize behavioral non-redundancy, provided that the chosen representation and similarity function are reported~\citep{friedmanVendiScore2023}.

\paragraph{ACE-conditioned Coverage.}
Coverage should be measured after the accuracy gate and within declared complexity ranges. Otherwise, invalid samples may make a dataset appear broader, or most of its valid coverage may be concentrated among trivial tasks. Aggregate statistics can also hide low validity in rare domains or long trajectories. Reports should therefore pair diversity statistics with validity rates, failure types, and model-relative difficulty for each important factor slice.

\paragraph{Transfer Coverage.}
Internal coverage does not by itself show that the data support generalization. Stronger evidence comes from controlled evaluation under named held-out factors, such as unseen tools, schemas, states, policies, interfaces, user behaviors, dynamics, or horizons. Studies should report both performance on the held-out factor and its gap from a matched in-domain setting~\citep{lvAgentsGeneralizeOpen2026,wangMCPBenchBenchmarking2026,wuMCPMarkBenchmark2026,liComplexMCPEvaluation2026}.

\paragraph{Marginal Learner Utility.}
The final question is whether a newly covered region improves the learner more than adding the same amount of data from existing regions. A sample may be new to the corpus but already easy for the model, or novel to the model but too difficult or irrelevant to produce useful learning. Model-aware generators approximate this distinction with capability profiles or success-rate bands~\citep{guoGenEnv2025,kangLearningChallenges2026,chenAgentFrontier2025,wolfBreakingSolverBottleneck2026}. Evaluation should therefore distinguish corpus novelty, model-relative novelty, and downstream transfer benefit.

\subsection{Diversity Tradeoffs and Limitations}
\label{sec:diversity-tradeoffs}

\paragraph{Diversity versus Quantity.}
Structural coverage can provide greater marginal value than repeated sampling from saturated modes~\citep{chenDIVEScalingDiversity2026,chenBeyondQuantityTrajectory2026}. This does not make quantity irrelevant: additional samples improve estimation and robustness in genuinely under-sampled regions. The limitation is that raw counts cannot distinguish support expansion from denser repetition, so scaling claims require coverage-controlled comparisons.

\paragraph{Diversity versus Accuracy.}
Broader recombination increases the chance of incompatible tools, unreachable goals, inconsistent states, and misaligned verifiers. Execution, procedural tests, constraints, and state-based checks must therefore accompany expansion~\citep{liuAPIGenAutomatedPipeline2024,songEnvScalerScalingToolInteractive2026,tianASTRAAutomatedSynthesis2026,ivanovAnchorMitigatingArtifact2026}. Yet verifier support can be narrower than the valid solution space: optimizing admission against one fixed checker may discard legitimate strategies and make the accepted set appear less diverse. Diversity should expand only after validity is established, while verifier coverage and independent audits expand with it.

\paragraph{Diversity versus Model-aware Complexity.}
A broad dataset can still be poorly allocated for a particular learner. Easy regions may be diverse but provide little new signal, whereas frontier-only generation can discard foundational capabilities and repeatedly chase transient failures. Because the useful complexity region moves during training, generation needs both frontier-focused allocation and broad replay coverage. Diversity should thus be reported within difficulty bands, and adaptation should be evaluated for collapse across skills and domains.

\paragraph{Realism, Controllability, and Scale.}

Environment-free trajectory generation sharpens this tradeoff by scaling from API specifications alone while shifting accuracy assurance to simulator consistency and model-based filtering~\citep{liSimia2025,leeEnvironmentFree2026}.
Real systems expose authentic dynamics and long-tail behavior but are costly, unsafe, difficult to reset, and prone to temporal drift. Learned simulators scale cheaply but may invent transitions; programmatic and recreated environments offer control and deterministic checking but omit unmodeled behavior. Hybrid designs combine real specifications or traces with executable replicas and learned components~\citep{wangAgentWorldModel2026,chaeSafeScalableWeb2026,zuoQwenAgentWorldLanguageWorld2026}. Their diversity claims remain conditional on simulator fidelity and should be tested on held-out real environments.

\paragraph{Mixture Interference and Open-world Drift.}
Combining domains and generators introduces incompatible formats, action conventions, reward scales, and sampling frequencies. Standardization and balancing can reduce this interference~\citep{chenAgentFLANDesigningData2024,zhangAgentOhanaDesignUnified2024,caiAutoForgeAutomatedEnvironment2025}, but aggressive normalization may erase domain-specific semantics. Moreover, any finite mixture is a snapshot: tools, interfaces, policies, and user behavior continue to change. Sustainable diversity therefore requires provenance tracking, shift detection, and periodic support repair rather than one-time maximization~\citep{lvAgentsGeneralizeOpen2026,dongAgentWorldScalingRealWorld2026,gaoSelfEvolvingSyntheticData2026}.

\section{Discussion}
\label{sec:discussion}

The preceding sections use ACE to analyze the construction and selection of agentic data. Several important questions, however, concern choices around agentic data generation rather than another mechanism for improving Accuracy, Complexity, or Diversity. This section discusses such extensions.

\subsection{Scaling Law under ACE Objective}

Data scaling primarily asks whether performance improves as more training examples are added. Under the ACE objective, however, raw quantity becomes a weaker proxy for useful scale because additional samples contribute unevenly. Accuracy determines how much of the generated pool is actually admissible; Complexity determines whether valid samples remain informative for the current learner rather than being already saturated or completely beyond reach; and Diversity determines whether new data expand behavioral support or merely densify modes that are already well covered. Recent results in agentic settings already suggest this distinction: diversity-oriented scaling can outperform simple quantity scaling under comparable or even smaller data budgets, while environment-scaling studies show that a smaller set of robust, behaviorally distinct environments can be more effective than a much larger but redundant or weakly verified collection~\citep{chenDIVEScalingDiversity2026,chenBeyondQuantityTrajectory2026,xuEnvFactoryScalingToolUse2026,tuScaleEnvScalingEnvironment2026}.

This perspective suggests that the relevant scaling variable for agentic data is closer to \emph{effective support} than raw dataset size. Scaling can occur by producing more valid experience, moving probability mass toward the learner's useful complexity frontier, or expanding coverage over environments, states, tasks, policies, and interaction structures. These directions are coupled rather than independent~\citep{guoGenEnv2025,chenAgentFrontier2025,wolfBreakingSolverBottleneck2026,haoFailureMasteryGenerating2026}. The strongest scaling strategy may not maximize any single ACE dimension, but maintain a growing region that is simultaneously valid, learnable, and behaviorally non-redundant.

A further implication is that agentic data scaling is likely to become increasingly dynamic. As the learner improves, previously useful tasks become saturated, the complexity frontier moves, and previously diverse regions can become redundant; at the same time, expanding into new domains introduces new accuracy requirements and verifier gaps. Static quantity scaling therefore has diminishing value unless generation and allocation adapt with the learner. Self-evolving and difficulty-aware systems already move in this direction by using failures, success rates, and coverage gaps to decide what experience to generate next~\citep{zhaiAgentEvolverEfficientSelfEvolving2025,kangLearningChallenges2026,guoGenEnv2025,chenAgentFrontier2025}. From the ACE perspective, a future scaling law for agentic data may therefore be less about how performance grows with the number of trajectories, and more about how efficiently additional generation expands the accurate, appropriately challenging, and behaviorally distinct experience available to the agent.

\subsection{Real and Synthetic Data under ACE}

We distinguish real and synthetic data by the origin of their agentic content. \emph{Real data} are collected from naturally occurring or deployed environments, such as human working with AIs. Their environments, tasks, or behaviors originate from actual use rather than being created specifically by the generation pipeline. \emph{Synthetic data} are constructed by models, programs, simulators, or controlled transformations to instantiate new environments, tasks, interactions, or supervision signals.
They provide different kinds of evidence, and neither is uniformly better under ACE. 

For \emph{Accuracy}, real data offer direct evidence of authentic interfaces, behavior, and user needs, but they are not automatically correct. Logs can contain failed or inefficient behavior, hidden context, stale interfaces, privacy-sensitive content, and outcomes that cannot be reconstructed or verified. Synthetic generation can attach explicit states, constraints, and verifiers and can regenerate failed cases, but its correctness is limited by the fidelity of the generator and environment. Model-generated tasks may be plausible yet infeasible, or overly machine-like. Real sources are therefore strongest for grounding and external auditing; synthetic sources are strongest when controlled execution and explicit supervision are required~\citep{liuAPIGenAutomatedPipeline2024,trivediAppWorldControllable2024,xuEnvFactoryScalingToolUse2026,chaeSafeScalableWeb2026}.

For \emph{Complexity}, real workflows naturally contain coupled constraints, long-tail states, delayed effects, and organizational rules, but their difficulty is difficult to control and may fall outside a learner's useful range. Synthetic pipelines can vary dependencies, information access, horizon, and assistance to target a model-relative frontier, although nominally longer or more elaborate samples may still be artificial or irrelevant. For \emph{Diversity}, real ecosystems expose authentic variation across users, tools, repositories, and temporal conditions, yet observed data are often highly skewed toward frequent workflows. Synthetic generation can deliberately fill missing combinations and counterfactual cases at scale, but shared prompts, teachers, and templates can create data with narrow behavioral support~\citep{liuToolACEWinningPoints2025,guoGenEnv2025,chenDIVEScalingDiversity2026,chenBeyondQuantityTrajectory2026}.

These tradeoffs favor hybrid allocation rather than a fixed real-to-synthetic ratio. Real data can define deployment-relevant factors, seed environment semantics, and provide held-out audits; synthetic data can expand, balance, and calibrate the support around those anchors. The appropriate mixture depends on which ACE deficiency is limiting: additional real evidence is valuable when fidelity is uncertain, whereas controlled synthesis is valuable when verified coverage or model-aware difficulty is missing. 
% Comparisons should consequently match token and training budgets and report results by provenance, because aggregate gains cannot reveal whether synthesis broadened useful support or merely amplified an existing source distribution.

\subsection{Data Generation for Agentic Pre-training and Mid-training}

% Source composition is one boundary of the framework; the training stage and unit of supervision form another. 
Most agentic data generation currently serves post-training: SFT teaches an interaction format and RL optimizes behavior from environment feedback. This places a large burden on a relatively late and data-limited stage, because a general foundation model must acquire basic interaction priors while also learning how to solve particular tasks. Agentic pre-training and mid-training instead expose models earlier to state transitions, action--observation dependencies, tool composition, persistent policies, and long-horizon information seeking~\citep{zengDaVinciDev2026,luYoutuLLM2025,tongyiDeepResearch2025,zuoQwenAgentWorldLanguageWorld2026}. The goal is not to replace post-training, but to provide reusable concepts on which later alignment and policy optimization can build.

This shift changes both scale and data form. Complete expert rollouts in real environments are too expensive to supply pre-training-scale corpora, motivating executable or learned environments and the conversion of repositories, videos, documents, and relational structures into interaction-relevant supervision~\citep{chenSWEUniverse2026,luVideoAgentTrek2025,xiongVideo2GUI2026,zhouDeepResearchPretraining2026}. Moreover, a training unit need not be a complete $(E,q,\tau,v)$ record anymore. Local state transitions, inverse-dynamics examples, dependency completions, reachability objectives, and distilled policy or causal knowledge can teach components from which agentic behavior is later composed~\citep{liuVisualStateTransitions2026,leiState2State2026,wangFunctionAwareFIM2026,wuGUICIDER2026}.

ACE consequently changes emphasis. Accuracy remains a floor, but often concerns local transition correctness, source fidelity, and distribution-level noise rather than verified task completion. Complexity concerns the information and dependency burden of fragments, not only end-to-end task difficulty. Once scalable quality control is available, diversity becomes especially important because pre-/mid-training should establish broad priors before downstream tasks are known. Promising directions include using post-training feedback to select or weight earlier-stage data, combining strongly verified anchors with much larger weakly supervised corpora, and determining which interaction knowledge should be internalized rather than kept explicit and updateable. 
% Progress will require matched studies that separate the effects of agentic structure from additional domain tokens and compute.

\subsection{Data Generation for Self-Evolving Agents}

Self-evolving agents change data generation from an offline preparation step into part of a continual learning loop. Instead of constructing a fixed dataset before training, the system observes the agent's behavior, identifies capability gaps, generates or discovers relevant experience, and uses verified outcomes to guide the next update. Recent systems instantiate parts of this loop by generating tasks from model failures or curiosity, coevolving environments and world models, and extracting reusable skills from accumulated traces~\citep{zhaiAgentEvolverEfficientSelfEvolving2025,fangWebEvolver2025,fuSESA2026,xiaoSocraticSWE2026}. Data generation therefore becomes a mechanism for deciding what the agent should experience next, rather than merely producing more samples.

Its role also expands beyond synthesizing complete tasks and successful trajectories. A self-evolving pipeline may create targeted tasks around a failure, modify their difficulty, explore an environment for new reachable states, repair an invalid rollout, extract a skill from a successful one, or replay earlier experience to preserve existing capabilities. Model-aware generators already use rollout success to move tasks toward the learner's current frontier~\citep{guoGenEnv2025,chenAgentFrontier2025}, while self-evolving systems improve sample use through experience reuse, attribution, and trace-derived skill discovery~\citep{zhaiAgentEvolverEfficientSelfEvolving2025,xiaoSocraticSWE2026}. The central problem consequently shifts from generating each sample independently to selecting, transforming, and allocating experience according to its expected value for the current agent.

ACE becomes dynamic in this setting. Accuracy prevents erroneous feedback from being repeatedly reinforced; Complexity determines whether new experience should be made harder, simplified, or supported as the capability frontier moves; and Diversity prevents the loop from concentrating only on recent failures or verifier-friendly strategies. These objectives cannot be estimated solely from the evolving agent, because a generator, learner, and verifier that adapt together may confirm the same mistaken assumptions. Stable self-evolution therefore requires fixed or independently updated anchors, such as held-out tasks, external execution, or periodic evaluation in real environments. The open challenge is not simply autonomous data generation, but a feedback loop that continues to expand capability without amplifying errors, narrowing coverage, or forgetting previously learned behavior.

\section{Conclusion}

Agentic data generation is the joint construction of an actionable environment, a grounded task, an interaction process, and, when needed, a trustworthy success signal. This paper combines a factorized account of how these components are generated with the ACE account of how the resulting distribution should be shaped. The former connects forward, task-first, trajectory-first, and structure-first pipelines; the latter explains why producing plausible individual samples is insufficient for building useful agent data.

The literature supports an asymmetric interpretation of ACE. Accuracy establishes the feasible set through consistency among environments, tasks, interactions, and verifiers. Complexity should then be calibrated relative to the learner and execution configuration rather than maximized through superficial length or structure, while Diversity should measure valid behavioral coverage rather than sample count or surface variation. Together, these dimensions suggest evaluating generation pipelines by the reliable and non-redundant learning value they add, while making verification coverage, model dependence, and generation costs explicit.

The field is now moving from fixed post-training trajectories toward generated environments, broader pre- and mid-training supervision, and feedback-driven experience that evolves with the agent. In this setting, data generation becomes less a one-time production step and more a continual process of discovering capability gaps, constructing and verifying relevant experience, and allocating it as the learner changes. The central challenge is therefore not autonomous generation alone, but maintaining a grounded learning loop that expands capability without amplifying errors, narrowing coverage, or losing contact with real environments.

% \newpage

\bibliographystyle{plainnat}
\bibliography{references_normalized}

\end{document}